\documentclass{article}

\usepackage{arxiv}

\usepackage[utf8]{inputenc} 
\usepackage[T1]{fontenc}    
\usepackage{hyperref}       
\usepackage{url}            
\usepackage{booktabs}       
\usepackage{amsfonts}       
\usepackage{nicefrac}       
\usepackage{microtype}      
\usepackage{lipsum}
\usepackage{graphicx}
\graphicspath{ {./images/} }

\usepackage{kpfonts}
\usepackage{siunitx}
\usepackage{caption}
\usepackage{subcaption}
\usepackage{lineno}

\def\gaheteSone{Table S1}
\def\gaheteStwo{Table S2}
\def\gaheteSthree{Figure S3}
\def\gaheteSfour{Figure S4}
\def\gaheteSfive{Figure S5}
\def\gaheteSsix{Figure S6}
\def\gaheteSseven{Figure S7}
\def\gaheteSeight{Table S8}
\def\gaheteSnine{Table S9}
\def\gaheteSten{Table S10}
\def\gaheteSeleven{Table S11}

\title{Genetic heterogeneity analysis using genetic algorithm and network science}

\author{
 Zhendong Sha \\
  School of Computing\\
  Queen's University\\
  Kingston, Ontario, K7L 2N8 \\ Canada \\
  \texttt{zhendong.sha@queensu.ca} \\
   \And
 Yuanzhu Chen \\
  School of Computing\\
  Queen's University\\
  Kingston, Ontario, K7L 2N8 \\ Canada \\
  \texttt{yuanzhu.chen@queensu.ca} \\
  \And
 Ting Hu \\
  School of Computing\\
  Queen's University\\
  Kingston, Ontario, K7L 2N8 \\ Canada \\
  \texttt{ting.hu@queensu.ca} \\
}

\begin{document}
\maketitle
\begin{abstract}
Through genome-wide association studies (GWAS), 
disease susceptible genetic variables 
can be identified by 
comparing the genetic data of individuals 
with and without a specific disease. 
However, the discovery of these associations
poses a significant challenge 
due to genetic heterogeneity and feature interactions. 
Genetic variables intertwined with these effects 
often exhibit lower effect-size, 
and thus can be difficult to be detected using machine learning feature selection methods.
To address these challenges, 
this paper introduces a novel feature selection mechanism for GWAS, 
named \textbf{F}eature \textbf{C}o-\textbf{s}election \textbf{Net}work (FCS-Net). 
FCS-Net is designed to extract 
heterogeneous subsets of genetic variables 
from a network 
constructed from multiple independent feature selection runs 
based on a genetic algorithm (GA),
a evolutionary learning algorithm.
We employ a non-linear machine learning algorithm 
to detect feature interaction.
We introduce the \textit{Community Risk Score} (CRS), 
a synthetic feature designed to quantify the collective disease association 
of each variable subset.
Our experiment showcases 
the effectiveness of the utilized GA-based feature selection method 
in identifying feature interactions
through synthetic data analysis. 
Furthermore, we apply our novel approach
to a case-control colorectal cancer GWAS dataset.
The resulting synthetic features 
are then used to explain the genetic heterogeneity 
in an additional case-only GWAS dataset. 
\end{abstract}

\keywords{Colorectal cancer \and Epistasis \and Genome-wide association studies \and Heterogeneity analysis \and Feature selection \and Genetic algorithm \and Network science}


\section{Introduction}
The primary objective 
of case-control genome-wide association studies (GWAS) 
is the identification of disease-associated genetic variables~\cite{rood2021legacy}. 
Over the past two decades, 
GWAS has successfully linked 
thousands of genetic variables to disease susceptibility~\cite{visscher2021discovery,visscher201710,loos202015}. 
However, understanding the genetic underpinnings of disease 
and the identification of high-risk patients 
remains highly complex~\cite{dahl2020genetic}. 
This complexity originates from multiple sources. 
First, genetic variables interact in a polygenic manner. 
Second, different genetic variables may interact with each other, 
a phenomenon known as epistasis~\cite{wray2018common}.
Finally, similar to the hidden class problem in machine learning~\cite{urbanowicz2013role}, disease-associated genetic variables 
may exhibit heterogeneity across different individuals~\cite{dahl2020genetic}. 
The combined impact of these factors 
will diminish the effect size of disease-related variables, 
thus making their identification challenging.

The identification of heterogeneity
is essential for the clinical translation of GWAS results. For instance, individuals who carry the {\tt BRA1}/{\tt BRA2} gene, 
constituting about 1\% of the population, 
exhibit a lifetime risk of developing breast cancer at 65\%/45\%, 
which is substantially higher than 
the population average of 12\%~\cite{torkamani2018personal}. 
For this cohort, specialized prevention measures 
are recommended to improve treatment outcomes 
through early detection of disease onset~\cite{Gabai-Kapara14205}.

Heterogeneity analysis is notably difficult in computational genetics 
due to the polygenic nature of genetic data
and epistasis. 
The detection of risk associations can be challenging 
if patients of a particular subtype 
are underrepresented in the population~\cite{urbanowicz2013role}. 
A simulation study has shown that 
genetic heterogeneity can 
prevent the detection of genetic associations 
by multifactor dimensionality reduction (MDR), 
a software frequently employed to capture genetic interactions~\cite{ritchie2003power}. 
Therefore, prior to MDR analysis, 
it can be helpful to identify 
clusters of individuals with similar genetic backgrounds~\cite{ritchie2003power}.

\begin{figure}
  \centering
  \includegraphics[width=\linewidth]{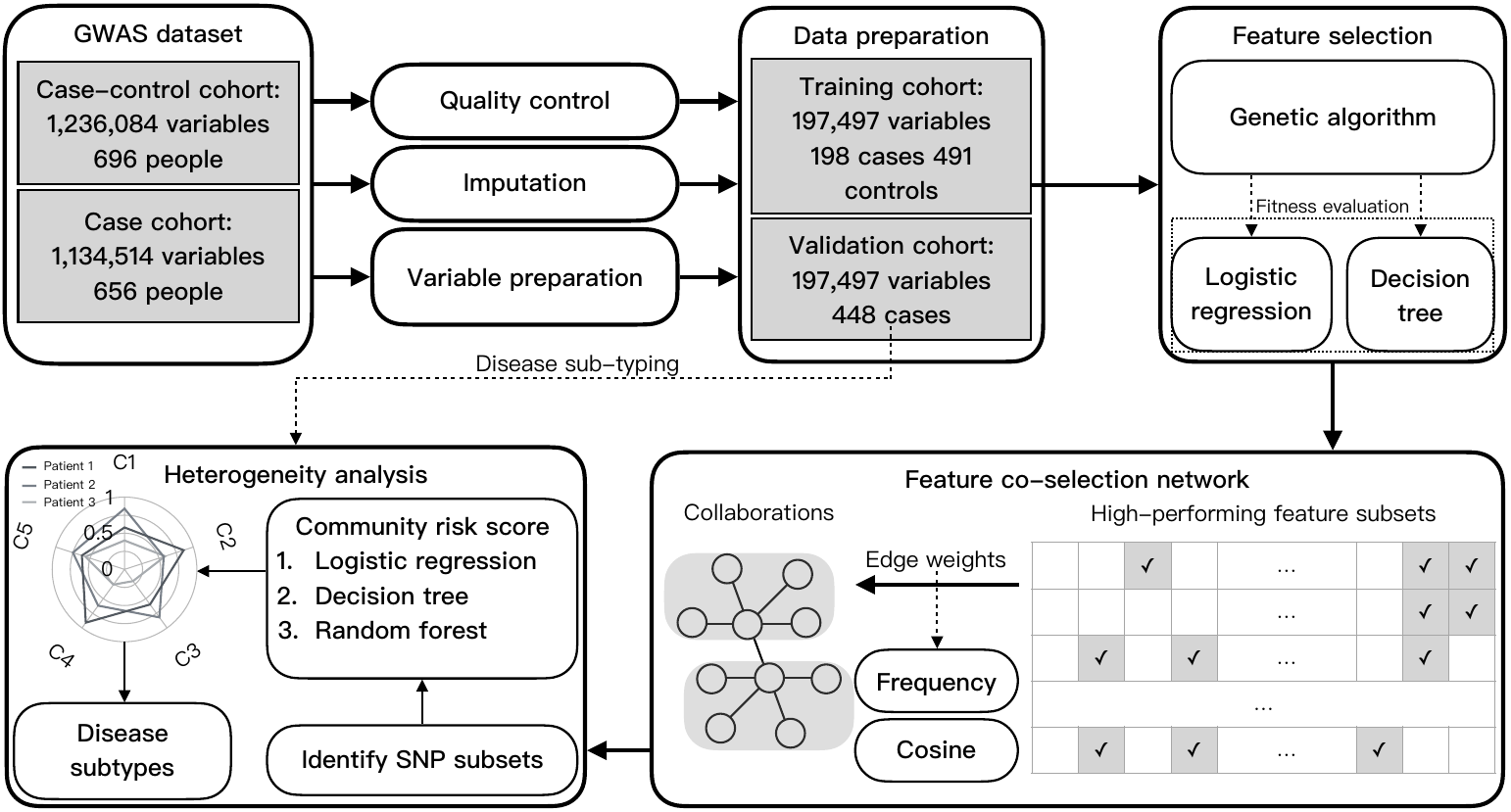}
  \caption[Overview of the FCS-Net framework]{Overview of the FCS-Net framework. The GA-based feature selection algorithm is repeated multiple times to collect a set of high-performing feature subsets. A network of genetic variables is then constructed to capture co-selected feature collaborations. We apply a community detection algorithm to find frequently co-selected groups of features. These groups are used to create new synthetic features (community risk scores) to identify disease subtypes.}
  \label{fig:flowchart}
\end{figure}

An effective heterogeneity analysis should address polygenicity and feature interaction together.
This study introduces a novel approach, 
named \textbf{F}eature \textbf{C}o-\textbf{s}election \textbf{Net}work (FCS-Net) (depicted in Figure~\ref{fig:flowchart}), 
to characterize genetic heterogeneity.
We frame the task of identifying heterogeneous feature subsets 
into a combinatorial optimization problem. 
We expect that multiple feature selection runs
will yield diverse feature subsets 
when applied to a heterogeneous dataset. 
The feature selection algorithm 
is iterated multiple times 
to gather a set of high-performing feature subsets. 
A network of genetic variables 
is constructed to capture co-selected feature collaborations. 
A community detection algorithm 
is then leveraged to identify closely collaborated feature communities, 
which we interpret as evidence of heterogeneous disease associations within the dataset. 
Synthetic features reflecting the collective disease association of each variable community are created
and hierarchical clustering algorithm is utilized 
to identify disease subtypes among diseased individuals 
based on these synthetic features. 
To confirm the capability of feature selection algorithms 
in identifying feature interactions, 
we undertake a simulation study based on GAMETES~\cite{urbanowicz2012gametes}, a software package capable of generating intricate biallelic SNP disease models.
We adopt the proposed framework 
on a case-control colorectal cancer GWAS dataset 
and carry out disease subtyping 
for another colorectal cancer GWAS dataset 
composed solely of diseased individuals.

\section{Backgrounds}

This section covers existing genetic heterogeneity analysis and feature selection methods. Feature selection plays a crucial role in identifying disease subtypes and mitigating the curse of dimensionality in high-dimensional datasets.

Clustering analysis has been employed 
to detect genetic heterogeneity 
in case-control studies~\cite{schork200114}. 
Traditional clustering techniques 
such as {\it k}-Means~\cite{li2018heterogeneity}, 
hierarchical clustering~\cite{terao2019distinct}, 
and topological data analysis~\cite{hinks2016multidimensional,li2015identification,nicolau2011topology}, 
have been leveraged to identify disease subtypes~\cite{dahl2020genetic}. 
Nevertheless, these techniques 
do not scale well with high-dimensional data. 
The learning classifier system (LCS), in contrast, 
deviates from the standard single model approach
by evolving a solution comprising multiple rules of clustering~\cite{urbanowicz2010application}. 
For instance, the extended Supervised Tracking and Classifying System (ExSTraCS) 
incorporates a feature-tracking strategy 
to identify features that have significant contributions 
to the accurate prediction of each instance~\cite{urbanowicz2012instance,urbanowicz2018attribute}.
Clustering feature-tracking scores 
thus allows for the detection of underlying heterogeneous subtypes~\cite{urbanowicz2013role,urbanowicz2012instance}. 
The LCS Discovery and Visualization Environment (LCS-DIVE)~\cite{zhang2021lcs} 
is an automated LCS model interpretation pipeline 
for biomedical data classification, 
LCS-DIVE utilizes feature-tracking scores and/or rules 
to characterize underlying heterogeneity through clustering.

For high-dimensional datasets, 
feature selection can mitigate the curse of dimensionality 
by discarding single nucleotide polymorphisms (SNPs) or genetic variables 
that are apparently not associated with the disease outcome. 
Existing literature groups feature selection methods 
into three categories: filter, wrapper, and embedded approaches~\cite{dash1997feature,guyon2003introduction}. 
The filter approach typically operates independently of the classification algorithm. 
In contrast, the wrapper approach employs a classification model 
to evaluate the quality of a feature subset, 
potentially enhancing the predictive power of the selected feature subset~\cite{dash1997feature,Liu2009,liu2010feature}.
However, the wrapper approach requires 
more computational resources, given its iterative evaluation of feature subsets,
compared to the filter approach. 
The embedded approach concurrently 
performs feature selection 
and predictive model construction. 
Random forest~\cite{breiman2001random} is a frequently used embedded feature selection algorithm~\cite{dash1997feature,Liu2009,liu2010feature}. 
Regularization techniques, such as Lasso or Ridge regression~\cite{Mak2017}, can automatically select the most important features by integrating the loss function to penalize the inclusion of less important features.
Current computational genetics research 
typically relies on filter approaches 
such as Relief-based methods~\cite{kira1992practical,urbanowicz2018relief,urbanowicz2018benchmarking,yang2008feature} and linkage disequilibrium (LD), 
i.e., the correlation structure among genetic variables~\cite{visscher201710,vilhjalmsson2015modeling}. 
Relief-based methods utilize distance measures to estimate the importance of each feature 
but do not distinguish if the importance arises 
from uni-variable disease-association or interactions with other features~\cite{urbanowicz2018relief}.

One type of wrapper feature-selection approaches 
employ evolutionary algorithms 
to search for the most relevant and influential feature subset. 
Evolutionary algorithm is a population-based optimization strategy that simulates natural evolution. 
Evolutionary algorithms 
like genetic algorithm (GA)\cite{siedlecki1993note,al2017examining,swerhun2020summary,sayed2019nested,garcia2020unsupervised}, 
particle swarm optimization (PSO)\cite{yang_pbmdr_2019}, 
and differential evolution (DE)\cite{Wang2022} 
have been utilized to perform feature selection on biological microarray datasets. 
For GA, he evolution strategy for the population of solutions consists of two components. Crossover involves combining multiple solutions to create new solutions. Mutation introduces random changes to individual solutions to create new solutions.
Specifically, GA searches for feature subsets 
by combining complementary subsets through crossover 
and making adjustments via mutation. 
A recent comparative study suggests 
that GA outperforms PSO 
on high-dimensional benchmark datasets from the UCI repository~\cite{nurhayati_particle_2020}. 
As a wrapper approach, 
GA-based feature selection relies on machine learning algorithms, such as linear regression~\cite{leardi1992genetic}, logistic regression~\cite{Sha2021}, Naive Bayes~\cite{da2011improving,canuto2012genetic,sousa2013email}, support vector machines (SVM)\cite{sayed2019nested,canuto2012genetic,seo2014feature}, and artificial neural networks (ANN)\cite{winkler2011identification,souza2011co,oreski2014genetic}, 
to evaluate the quality of a feature subset.

\section{Methodologies}

Polygenic disease associations and epistasis are major obstacles in identifying genetic heterogeneity. 
Our study addresses these issues by considering the collaborations of features in subsets derived from multiple feature selection runs (Section~\ref{sec:collab}). 
Additionally, this approach employs a feature construction method (Section~\ref{sec:featueSyn}) to capture the heterogeneous risk effects of different genetic variable subsets. 
The overview of our methodology is illustrated in Figure~\ref{fig:flowchart}.

\subsection{Colorectal cancer data}

The GWAS datasets studied in this research 
are from colorectal cancer transdisciplinary (CORECT) consortium
~\cite{ScOt15}.
Genotyping is conducted using a custom Affymetrix genome-wide platform (the Axiom CORECT Set) 
on two physical genotyping chips (pegs)~\cite{ScOt15}. 
A total of 696 samples (200 colorectal cancer cases and 496 controls) 
are genotyped using the first chip, 
with each sample comprised of 1,236,084 SNPs. 
A total of 656 cases are genotyped using the second chip, 
with each sample comprised of 1,134,514 SNPs. 
The data processing procedure consists of three parts. The first part is the pre-imputation process, which is followed by the imputation of the missing values using the Michigan Imputation Server (MIS)~\cite{das2016next}. The final step is the post-imputation process.

We perform quality control using PLINK~\cite{PuOt07}, 
a whole-genome association analysis software tool.
The pre-imputation process performs sample-level quality control to remove 
samples with a genotyping call rate less than 95\%, 
sex labelling is not consistent with the chromosome, 
sample heterozygosity is not within three standard deviations from the mean.
The pre-imputation process also perform SNP-level quality control to remove SNPs with minor allele frequency less than 1\%.
The two cohorts, genotyped using different chips, are merged and prepared to meet the requirements of MIS following the guidelines provided by its official tutorial\footnote{https://imputationserver.readthedocs.io/en/latest/prepare-your-data/}.

During the imputation step,
the Michigan Imputation Server is configured to use the eagle phasing algorithm~\cite{loh2016reference}, the hg19 reference panel and the mixed population option to accommodate the multi-racial population structure in Canada. 
The post-imputation process excludes low-quality SNPs based on the imputation R2 of minimac3 (R2>0.3)~\cite{das2016next}. 

We extract all SNPs of the dataset used for MIS submission based on chromosome position and perform minor allele frequency and linkage disequilibrium filtering ($r^2$ = 0.2). 
A SNP is removed if its minor allele frequency is less than 0.01. 
IBD, or Identity by Descent, in PLINK 
is a computational method used to estimate the proportion of the genome 
where two individuals share alleles from a common ancestor.
We remove first-degree relatives based on IBD. 
Samples in the second dataset 
are removed if the PI\_HAT value is above 0.5 
with any samples in the first dataset.

Finally, a total of 197,497 SNPs are selected for the subsequent analysis. 
The first dataset (training cohort) with both diseased cases (N=198) and healthy controls (N=491) is used for training, 
and the second dataset (validation cohort) is used for genetic heterogeneity analysis (N=448).

\subsection{Genetic algorithm}
\label{sec:ga_featureSel}

\begin{figure}
    \centering
    \includegraphics[width=.5\linewidth]{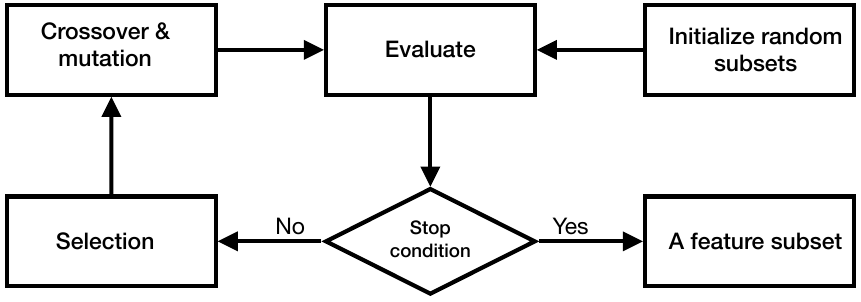}
    \caption{Flowchart of the GA algorithm}
    \label{fig:ga_flowchart}
\end{figure}

We employ genetic algorithm (GA) 
as the foundational search strategy for feature selection (Figure~\ref{fig:ga_flowchart}). 
In this approach, GA represents a feature subset as an individual for population-based evolution, 
which is a binary vector 
with a length identical to the total number of features in the data. 
Here, a ``1'' suggests the selection of the corresponding feature, 
while a ``0'' indicates its exclusion. 
Each individual (selected feature subset) has a fitness value 
that indicates its overall relevance to the disease prediction.
GA leverages tournament selection 
to select the best performing subsets from the population
and utilizes uniform crossover with a probability of $\textit{cxpb}$ 
and bit-flip mutation with a probability of $\textit{mutpb}$ 
to evolve these individuals.

GA employs a machine learning algorithm
to evaluate the fitness of its individual (feature subset).
Fitness is assessed as the average testing area under the Receiver Operating Characteristic curve (AUC-ROC) 
from a five-fold cross-validation~\cite{hanley1982meaning}. 
AUC-ROC is a key performance metric 
used to evaluate a predictive model's ability to distinguish between classes. 
We use two machine learning algorithms 
provided by {\tt scikit-learn}\cite{scikit-learn} 
to evaluate the fitness of a feature subset. 
The first algorithm is logistic regression\cite{cox1958regression}, 
due to it is common adoption in polygenic risk prediction models.
The second algorithm is the decision tree~\cite{breiman2017classification}, 
which characterizes the interactions between features on the prediction.
In Section~\ref{sec:synData_analysis}, 
the difference between these two algorithms 
is investigated using synthetic data.

To address the high-dimensional nature of GWAS data, 
we implement two modifications to the GA algorithm.
We introduce a size limit parameter, $\textit{size\_limit}$, 
which sets an upper limit on the number of features 
that can be selected in a feature subset. 
If mutation or crossover results in a selection exceeding the $\textit{size\_limit}$, 
we randomly deselect some features. 
This parameter aids in preventing overfitting 
by limiting the number of selected features. 
We have also noted that the crossover's efficiency 
is undermined by the fact that 
decision tree only use a fraction of the features 
in the feature subset. 
Consequently, the decision tree-based GA 
only performs crossover on features included in the decision tree and deselects the unused ones. 
Parameter configurations for all experiments in this study 
are provided in Table~\ref{tab:parameter_config}.

The open-source package Distributed Evolutionary Algorithm in Python (DEAP)~\cite{DEAP_JMLR2012} 
is used to implement the GA. 
The population size is defined as $\textit{pop\_size}$, 
and the maximum number of generations is $\textit{ngen}$. 
Tournament selection with $\textit{tour\_size}$ 
is used to choose parents for crossover and mutation. 
Uniform crossover with a probability $\textit{cxpb}$ and bit-flip mutation with a probability $\textit{mutpb}$ are utilized. 

\subsection{Feature collaboration identification}
\label{sec:collab}

Given the stochastic nature of GA-based feature selection, 
each feature selection iteration yields 
different evolved feature subsets. 
Therefore, we conduct multiple GA runs and 
assume that the disease association of a feature pair (or their collaboration) 
can be reflected by their co-selection frequency across independent feature selection runs.

We assemble a set $\Gamma$ 
of high-performing feature subsets 
derived from multiple GA-based feature selection iterations. 
A feature selection matrix $M$ 
is constructed for $\Gamma$ 
such that $M_{(i,j)}=1$ if feature $j$ is selected by subset $i$, 
and $M_{(i,j)}=0$ otherwise. 
To quantify the degree of feature collaboration, 
we use the co-selection frequency and pairwise cosine similarity. 
The co-selection frequency 
counts the instances when a pair of features are co-selected, 
while the cosine similarity measures the selection similarity 
between features across different subsets.

We compute a feature co-selection frequency matrix $A$ as follows:
\begin{equation}
\label{eq:adj}
A = M^\text{T}M,
\end{equation}
where each element $A_{(i,j)}$ in $A$ 
represents the number of evolved subsets in $\Gamma$ 
that select both features $i$ and $j$.
One caveat of frequency-based identification 
is its inability to distinguish between 
collaborations involving frequently selected features 
and those that involve interdependent features crucial for predictive performance. 
To mitigate this limitation, 
we additionally employ cosine similarity 
to measure the likeness between pairs of features 
selected by different feature subsets.

The cosine similarity $S_C$ of a feature pair $(A, B)$ 
is computed as the cosine similarity 
between any two columns of the matrix $M$:
\begin{equation}
    S_C(\Vec{A},\Vec{B}) = \frac{\sum_{i=1}^n A_i B_i}{\sqrt{\sum_{i=1}^n A_i^2}\sqrt{\sum_{i=1}^n B_i^2}}
\end{equation}
where $A_i$ and $B_i$ are the elements of column vectors $\Vec{A}$ and $\Vec{B}$ corresponding to subset $i$.

\subsection{Feature co-selection network}
\label{sec:cosel}

The feature co-selection network, 
denoted as $G_{\text{coSel}}=(N, E)$~\cite{Sha2021}, 
is designed to capture strong collaborations 
between feature pairs within $\Gamma$. 
In this network, each node $n_i \in N$ 
represents a feature (genetic variable), 
and each edge $(n_i,n_j) \in E$ 
represents the collaboration between the corresponding variables.

To include the most important pair-wise feature collaborations, 
we impose edge weight thresholding. 
Specifically, we eliminate all edges in $G_{\text{coSel}}$ 
with weights falling below $\tau_{\text{occ}}$ 
or $\tau_{\text{cos}}$, 
where $\tau_{\text{occ}}$ and $\tau_{\text{cos}}$ 
denote the edge thresholds for co-selection frequency 
and cosine similarity, respectively.

To determine the values of $\tau_{\text{occ}}$ and $\tau_{\text{cos}}$, 
we employ a network community detection algorithm (specifically, greedy~\cite{clauset2004finding}) on $G_{\text{coSel}}$ 
and measure the quality of community separations using network modularity~\cite{clauset2004finding}. 
A network with high modularity will have dense intra-community node connections 
and sparse inter-community node connections. 
We optimize network modularity 
since we assume that collaborations 
between genetic variables exist in a modular form.

\subsection{Community risk score}
\label{sec:featueSyn}

The Community Risk Score (CRS) 
aggregates the collective disease risk associated 
with a particular subset of genetic variables, 
equivalent to the network communities 
within the feature co-selection network. 
CRS can be used to estimate an individual's disease risk.

To compute CRS, we randomly select 80\% of the samples from the training cohort, 
which are then used to construct a predictive model 
for each variable subset corresponding to each CRS. 
This process will be repeated multiple times.
After conducting multiple resampling runs, 
the CRS value for each individual 
is the averaged probabilistic outputs 
across multiple resampling runs.

In this study, we use the CRS value 
to estimate an individual's disease risk 
with regard to a list of genetic variables. 
We use a variety of machine learning algorithms in parallel, 
including logistic regression, decision tree, and random forest, 
and perform 1000 resampling runs. 
The synthetic features generated by different communities are described using ``C'' 
followed by the community number. 
For example, the CRS value created for network community 1 is be named C1.

\subsection{Heterogeneity analysis using synthetic features}
\label{sec:hete}

We perform clustering analysis 
to reveal disease heterogeneity within the validation cohort. 
Initially, we represent each observation using CRS values.
This is followed by conducting hierarchical clustering 
using the Euclidean distance and the ward.D algorithm 
to group patients~\cite{kaufman2009finding}. 
JASP~\cite{JASP2019}, a platform-agnostic statistical software,
facilitates the clustering analyses. 
The derived clusters are visualized using cluster means plots.

\subsection{Functional enrichment analysis}
\label{sec:funcEnrich}

We employ functional enrichment analysis 
to gain insight into a list of genetic variables,
which is often used to translate disease-associated variables 
generated from genetic association analysis 
into biological insights. 
We use g:Profiler~\cite{gProfiler2019}, 
a frequently updated software, 
to perform enrichment analysis. 
The software includes many popular biochemical pathway databases, 
such as Gene Ontology (GO), molecular function (GO:MF), biological process (GO:BP), cellular component (GO:CC), and common biological data sources such as Kyoto Encyclopedia of Genes and Genomes (KEGG), Reactome (REAC), WikiPathways (WP), Transfac (TF), Human Protein Atlas (HPA), CORUM protein complexes (CORUM), and Human Phenotype Ontology (HP). 
The statistical significance of functional enrichment terms 
is determined using the well-proven cumulative hypergeometric test. 
Terms surpassing a $p$-value threshold of 5\% are considered as significant.

To explain the enrichment analysis derived from g:profiler, 
we use the procedure described in~\cite{reimand2019pathway}. 
EnrichmentMap~\cite{Merico2010} and AutoAnnotate~\cite{Kucera2016} 
are two popular enrichment analysis tools often used together. 
EnrichmentMap allows the visualization of enrichment results 
as a network of biological terms and comparison of enrichment results of different gene sets. 
In EnrichmentMap, nodes represent biological terms, and edges represent shared genes between two terms. 
AutoAnnotate identifies communities of connected terms and generates the theme of each community. 
Together, these tools help us to understand the functions of genes of interest.

\section{Results}
\subsection{Simulation study}
\label{sec:synData_analysis}
We conduct a simulation study 
to assess the efficacy of fitness evaluations in the GA
based on logistic regression and decision tree algorithms 
to identify phenotype associated features and interactions. 
The simulation study is conducted 
using an open-access dataset from the PMLB~\cite{Romano2021}, 
which is generated using the GAMETES tool~\cite{urbanowicz2012gametes}. 
GAMETES is frequently used in genetic studies to benchmark machine learning algorithms. 
The selected dataset\footnote{GAMETES\_Epistasis\_2\_Way\_1000atts\_0.4H\_EDM\_1\_EDM\_1\_1}
encompasses 1,000 attributes across two classes, 
with two attributes exhibiting pure epistatic interaction. 
The objective of the feature selection process 
is to pinpoint these two interacting attributes.

For each fitness evaluation method (logistic regression or decision tree), 
the GA is run for 1,000 iterations. 
The effectiveness of fitness measures 
is evaluated by the frequency at which the two epistatic attributes are included 
in the best-performing feature subset evolved by the algorithm. 
A detailed description of the parameter configuration 
is provided in Table~\ref{tab:parameter_config}.

Our results show that 
when using the fitness evaluation based on the decision tree algorithm, 
the two epistatic features exhibit a co-occurrence frequency of 19 
and a cosine similarity of 0.951. 
Conversely, using the logistic regression algorithm for fitness evaluation 
results in the GA's failure 
to identify the feature pair, 
as one of the features is never selected. 
This finding underscores the considerable influence of the choice of fitness evaluation algorithm 
on the identification of interacting feature pairs. 
In subsequent sections, 
we perform heterogeneity analysis 
using both fitness measures 
to get a comprehensive understanding of the features (genetic variables) and their interactions in the CRC GWAS data (See Section~\ref{sec:GA_DT} and Section~\ref{sec:GA_LR}).

\begin{table*}[]
\caption[The parameter configurations for genetic algorithm]{The parameter configurations for genetic algorithm.}
\label{tab:parameter_config}
\begin{tabular}{lllllll}
\toprule
Parameters            && Simulation study    &               && GWAS dataset        &               \\ \cline{1-1} \cline{3-4} \cline{6-7} 
Fitness     && Logistic regression & Decision tree && Logistic regression & Decision tree \\ \hline
$|\Gamma|$  && 1,000               & 1,000         && 10,000              & 10,000        \\
$\textit{pop\_size}$   && 200                 & 200           && 1,000               & 1,000         \\
$\textit{size\_limit}$ && 5                   & 5             && 200                 & 200           \\
$\textit{ngen}$      && 50                  & 50            && 50                  & 100           \\
$\textit{tour\_size}$  && 3                   & 3             && 6$^\ast$           & 6$^\ast$     \\
$\textit{cxpb}$        && 0.5                 & 0.5           && 0.8$^\ast$         & 0.8$^\ast$    \\
$\textit{mutpb}$       && 0.2                 & 0.2           && 0.2$^\ast$         & 0.2$^\ast$    \\
\bottomrule
\end{tabular}

$|\Gamma|$: Number of feature selection runs. 

$^\ast$: Determined through tuning three parameters, including $\textit{mutpb}=\{0.2,0.5,0.8\}$, $\textit{cxpb}=\{0.2,0.5,0.8\}$, and $\textit{tour\_size}=\{3,6,9\}$.
\end{table*}

\subsection{Heterogeneity analysis based on decision tree}
\label{sec:GA_DT}
This section presents the results 
of the heterogeneity analysis 
employing the decision tree algorithm as the fitness measure.
The parameter configuration for this analysis 
is provided in Table~\ref{tab:parameter_config}. 
Compared to logistic regression, 
the decision tree algorithm 
is capable of recognizing genetic interactions, 
which are considered to contribute more significantly to disease risk than individual variable effects~\cite{boyle2017expanded}. 
To highlight frequently co-selected feature pairs, 
we utilize cosine similarity (defined in Section~\ref{sec:cosel}) 
in conjunction with the co-occurrence-based metric 
to filter edges for the feature co-selection network $G_{\text{coSel}}^{\text{DT}}$.

We begin by outlining the process 
of constructing the co-selection network $G_{\text{coSel}}^{\text{DT}}$ 
based on the decision tree algorithm. 
We then investigate the heterogeneous risk associations of each synthetic feature 
and identify distinct disease subtypes. 
Lastly, we undertake genetic enrichment analysis 
to pinpoint the biological terms associated with colorectal cancer.

\subsubsection{The construction of the feature co-selection network}

\begin{figure}[t]
     \centering
     \begin{subfigure}[b]{0.45\textwidth}
         \centering
         \includegraphics[width=\textwidth]{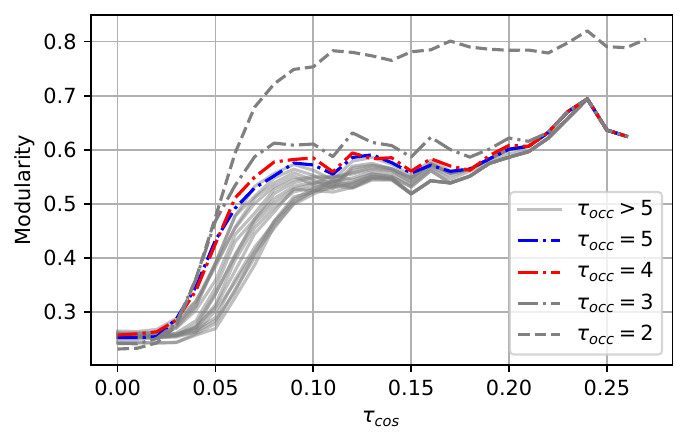}
         \caption{Determining the $G_{\text{coSel}}^{\text{DT}}$ network using feature co-selection number $\tau_{\text{occ}}$ and cosine similarity $\tau_{\text{cos}}$.}
        \label{fig:softthreshold_GADT}
     \end{subfigure}
     \hfill
     \begin{subfigure}[b]{0.45\textwidth}
         \centering
         \includegraphics[width=\textwidth]{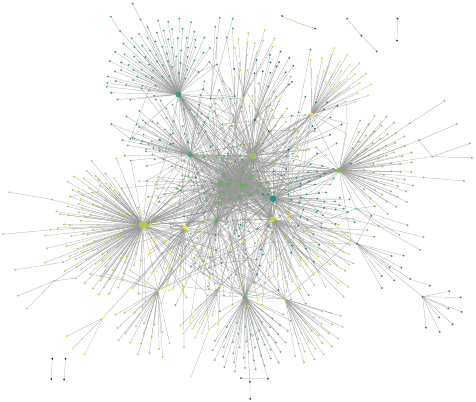}
         \caption{The $G_{\text{coSel}}^{\text{DT}}$ network generated using 10,000 feature selection runs based on decision tree with $\tau_{\text{occ}}=5$ and $\tau_{\text{cos}}=0.09$. }
  \label{fig:coSel_GADT}
     \end{subfigure}
        \caption[The determination and visualization of feature co-selection network based on decision tree]{The determination and visualization of feature co-selection network based on decision tree. We determine the $G_{\text{coSel}}^{\text{DT}}$ network using feature co-selection number $\tau_{\text{occ}}$ and cosine similarity $\tau_{\text{cos}}$. The $G_{\text{coSel}}^{\text{DT}}$ network is generated with $\tau_{\text{occ}}=5$ and $\tau_{\text{cos}}=0.09$. Network communities (N=20) are represented by colors. The size of node represents its degree.}
\end{figure}

The co-selection network $G_{\text{coSel}}^{\text{DT}}$ 
is created based on approximately 10,000 feature selection runs. The decision tree algorithm can identify feature interactions, 
and we leverage both cosine similarity and co-occurrence frequency 
to estimate the degree of interdependence 
between pairs of genetic variables (refer to Figure~\ref{fig:softthreshold_GADT}). 
Cosine similarity is used to detect variable pairs 
with fewer co-occurrences 
but exhibit substantial correlations 
across the evolved feature subsets.

The choice of edge weight threshold values 
is a crucial factor determining the structure of the network. 
As the $\tau_{\text{cos}}$ threshold increases, 
the measured network modularity increases, 
indicating a growth in the structural modularity of these features
generated by the decision tree based GA (see Figure~\ref{fig:softthreshold_GADT}).
Conversely, a high $\tau_{\text{occ}}$ threshold ($\tau_{\text{occ}}>5$) 
leads to network fragmentation rather than modularization, 
generating an exceedingly high number of disconected sub-networks (refer to Supplementary material {\gaheteSone}). 
Hence, despite its favorable network modularity metric, 
a high $\tau_{\text{occ}}$ threshold 
is inappropriate for feature selection analysis. 
The network modularity metric peaks 
when the $\tau_{\text{cos}}$ threshold exceeds 0.2, 
at which point the network 
consists of the most pertinent variable pairs 
with approximately 30 variables.

To balance network fragmentation 
and ensure the inclusion of sufficient genetic variables 
for our subsequent functional enrichment analysis, 
we establish the co-occurrence threshold $\tau_{\text{occ}}$ at 5 
and the cosine similarity cut-off $\tau_{\text{cos}}$ at 0.09 (where the network modularity equals 0.582). 
If the $\tau_{\text{cos}}$ threshold is elevated to 0.13 (network modularity equals 0.591), 
only 300 genetic variables remain within the network. 
This loss of genetic variables, compared to choosing $\tau_{\text{cos}}=0.09$ which contains 1036 variables, 
considerably constrains the effectiveness of the biological enrichment analysis.

\subsubsection{Heterogeneous predictive performance of community risk scores}
\label{sec:DTFR_Hete_CRS}

The constructed $G_{\text{coSel}}^{\text{DT}}$ network 
exhibits high network modularity. 
As depicted in Figure~\ref{fig:coSel_GADT}, 
the network comprises 20 communities of genetic variables, 
12 of which contain more than three genetic variables. 
In this section we investigate the heterogeneous predictive performance of synthetic features. 
Each synthetic feature 
corresponds to a distinct network community 
within the co-selection network.

A synthetic feature, named community risk score (CRS), 
estimates the risk impact of the genetic variables 
contained within each network community. 
For each community, we train 
logistic regression, decision tree, and random forest models
using the training cohort of data. 
The probabilistic output generated by these predictive algorithms 
will be employed to represent individuals in the validation cohort, 
as explained in Section~\ref{sec:featueSyn}.

\begin{figure}
     \centering
     \begin{subfigure}{0.19\textwidth}
         \centering
         \includegraphics[width=\textwidth]{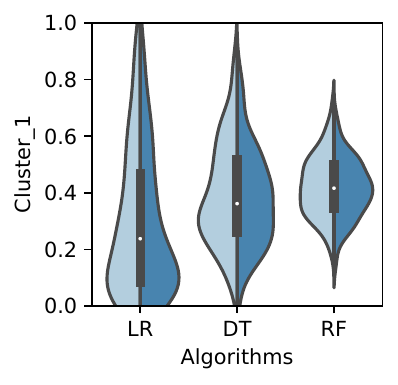}
         \label{fig:hete_gadt_c1}
     \end{subfigure}
     \begin{subfigure}{0.19\textwidth}
         \centering
         \includegraphics[width=\textwidth]{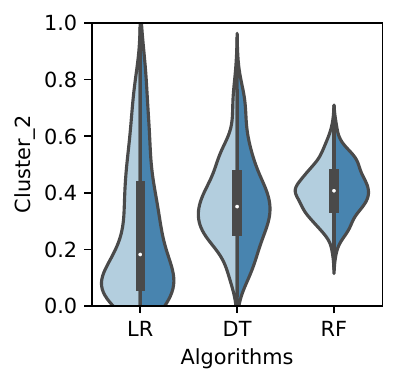}
         \label{fig:hete_gadt_c2}
     \end{subfigure}
     \begin{subfigure}{0.19\textwidth}
         \centering
         \includegraphics[width=\textwidth]{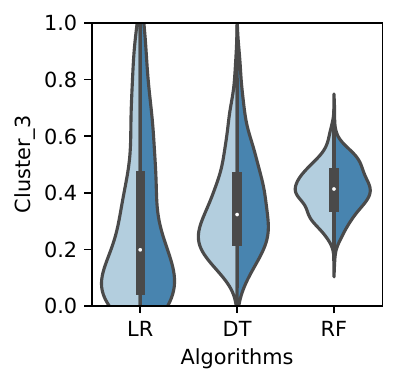}
         \label{fig:hete_gadt_c3}
     \end{subfigure}
     \begin{subfigure}{0.19\textwidth}
         \centering
         \includegraphics[width=\textwidth]{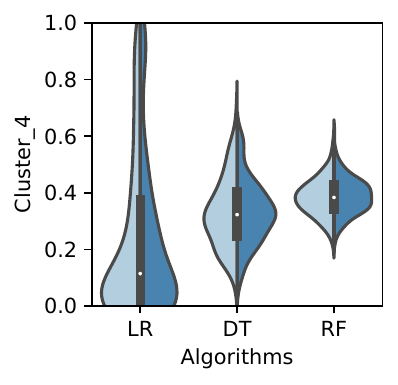}
         \label{fig:hete_gadt_c4}
     \end{subfigure}
     \begin{subfigure}{0.19\textwidth}
         \centering
         \includegraphics[width=\textwidth]{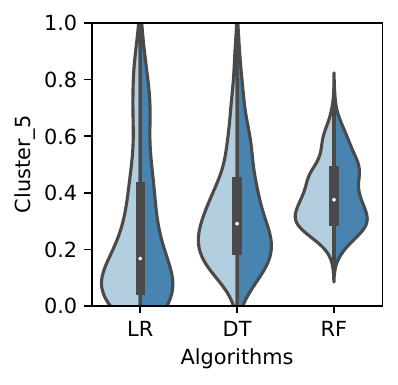}
         \label{fig:hete_gadt_c5}
     \end{subfigure}
     \begin{subfigure}{0.19\textwidth}
         \centering
         \includegraphics[width=\textwidth]{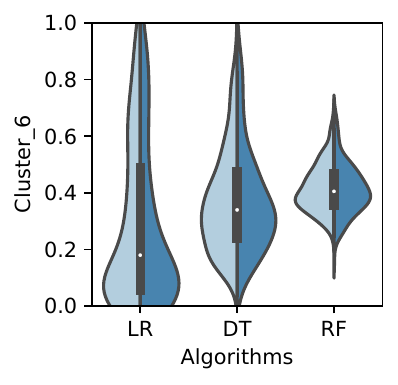}
         \label{fig:hete_gadt_c6}
     \end{subfigure}
     \begin{subfigure}{0.19\textwidth}
         \centering
         \includegraphics[width=\textwidth]{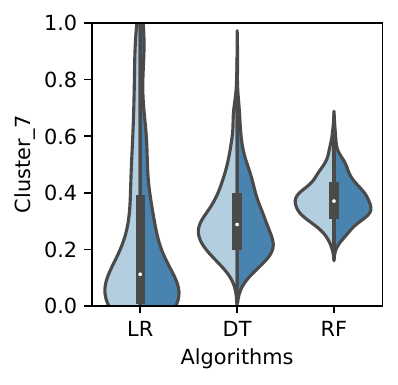}
         \label{fig:hete_gadt_c7}
     \end{subfigure}
     \begin{subfigure}{0.19\textwidth}
         \centering
         \includegraphics[width=\textwidth]{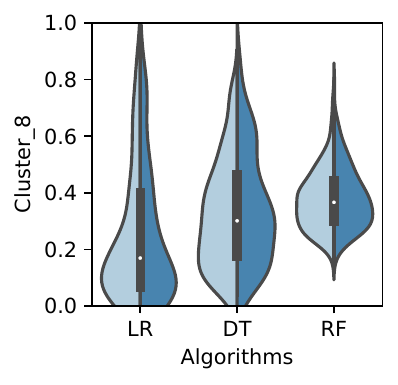}
         \label{fig:hete_gadt_c8}
     \end{subfigure}
     \begin{subfigure}{0.19\textwidth}
         \centering
         \includegraphics[width=\textwidth]{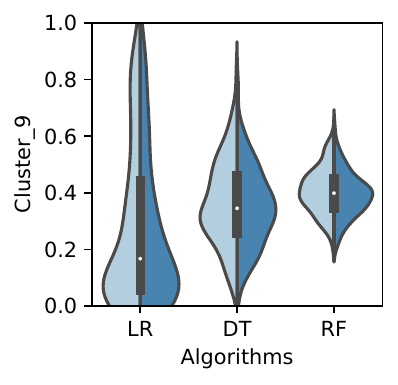}
         \label{fig:hete_gadt_c9}
     \end{subfigure}
     \begin{subfigure}{0.19\textwidth}
         \centering
         \includegraphics[width=\textwidth]{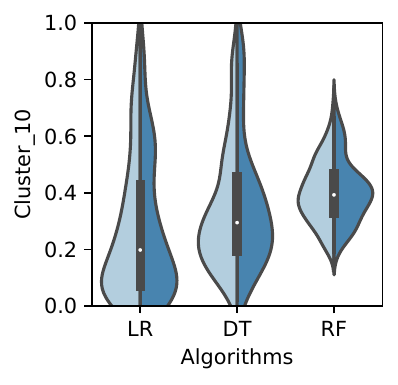}
         \label{fig:hete_gadt_c10}
     \end{subfigure}
     \begin{subfigure}{0.19\textwidth}
         \centering
         \includegraphics[width=\textwidth]{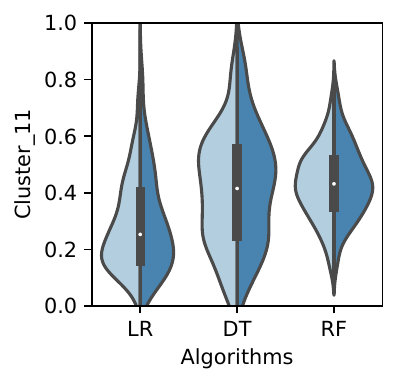}
         \label{fig:hete_gadt_c11}
     \end{subfigure}
     \begin{subfigure}{0.19\textwidth}
         \centering
         \includegraphics[width=\textwidth]{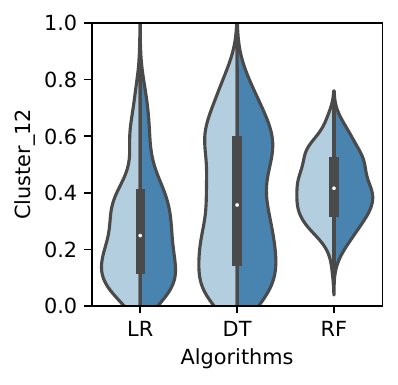}
         \label{fig:hete_gadt_c12}
     \end{subfigure}
     \begin{subfigure}{0.19\textwidth}
         \centering
         \includegraphics[width=\textwidth]{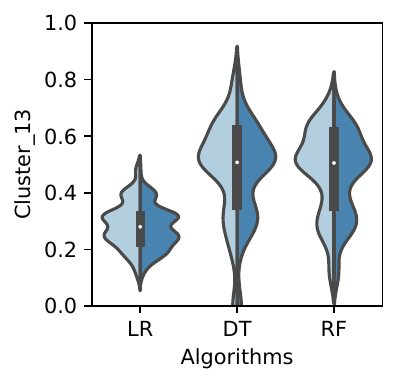}
         \label{fig:hete_gadt_c13}
     \end{subfigure}
     \begin{subfigure}{0.19\textwidth}
         \centering
         \includegraphics[width=\textwidth]{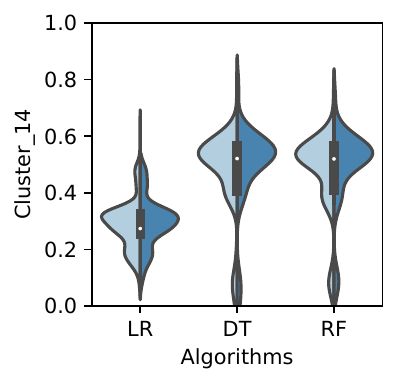}
         \label{fig:hete_gadt_c14}
     \end{subfigure}
     \begin{subfigure}{0.19\textwidth}
         \centering
         \includegraphics[width=\textwidth]{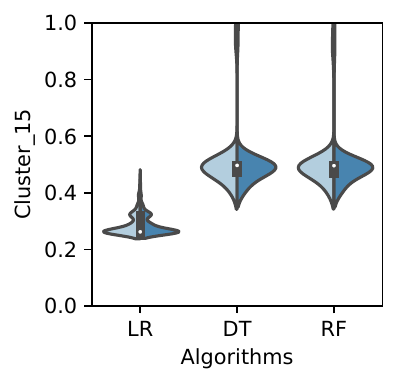}
         \label{fig:hete_gadt_c15}
     \end{subfigure}
     \begin{subfigure}{0.19\textwidth}
         \centering
         \includegraphics[width=\textwidth]{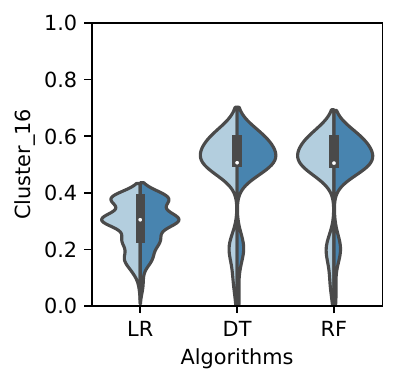}
         \label{fig:hete_gadt_c16}
     \end{subfigure}
     \begin{subfigure}{0.19\textwidth}
         \centering
         \includegraphics[width=\textwidth]{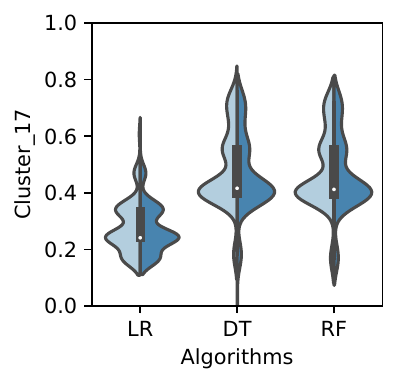}
         \label{fig:hete_gadt_c17}
     \end{subfigure}
     \begin{subfigure}{0.19\textwidth}
         \centering
         \includegraphics[width=\textwidth]{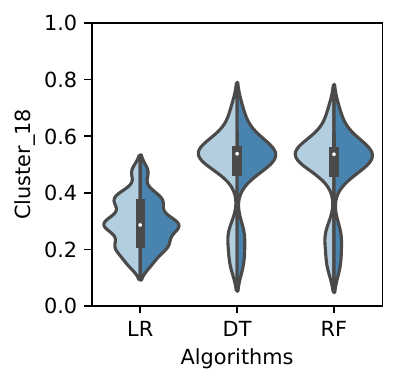}
         \label{fig:hete_gadt_c18}
     \end{subfigure}
     \begin{subfigure}{0.19\textwidth}
         \centering
         \includegraphics[width=\textwidth]{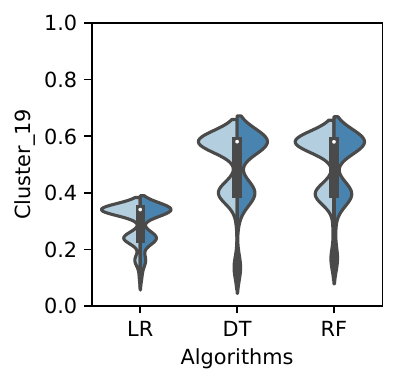}
         \label{fig:hete_gadt_c19}
     \end{subfigure}
     \begin{subfigure}{0.19\textwidth}
         \centering
         \includegraphics[width=\textwidth]{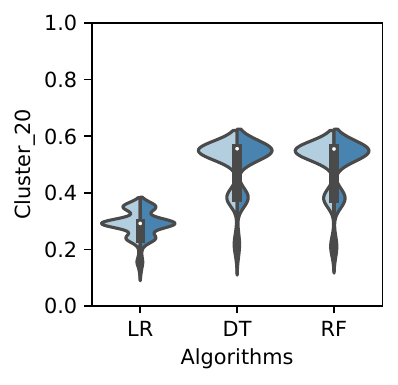}
         \label{fig:hete_gadt_c20}
     \end{subfigure}
     \begin{subfigure}{0.19\textwidth}
         \centering
         \includegraphics[width=\textwidth]{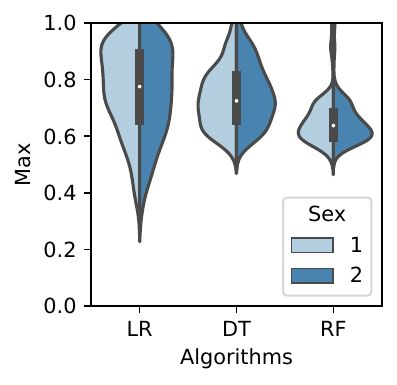}
         \label{fig:hete_gadt_max}
     \end{subfigure}
        \caption[The heterogeneous risk predictive performance of the CRSs based on decision tree]{The heterogeneous risk predictive performance of the CRSs based on decision tree. As many as 20 CRSs are identified from $G^{\text{DT}}_{\text{coSel}}$ and used to describe individuals in the validation cohort.The maximum risk impact of all communities is summarized in sub-figure entitled ``Max''.}
        \label{fig:hete_Pred_DT}
\end{figure}

As demonstrated in Figure~\ref{fig:hete_Pred_DT}, 
the logistic regression algorithm (labelled as LR), 
is unable to recognize the disease risk 
of observations within the validation cohort. 
This may stem from the inability of the linear regression-based model 
to capture the feature interactions present within the feature communities. 
Further exploration of the predictive performance of CRSs 
based on decision tree and random forest algorithms (labelled as DT and RF), 
exhibits improved results across all communities.
Lastly, although the CRSs based on the logistic regression algorithm exhibit weaker predictive performance 
than those based on the decision tree and random forest algorithms, 
the maximum risk impact across all 20 communities 
exhibits a better predictive power (refer to the sub-figure titled “Max” in Figure~\ref{fig:hete_Pred_DT}). 
This result suggests heterogeneity in the risk effects 
of the non-epistatic risk within the co-selection network $G_{\text{coSel}}^{\text{DT}}$.

\subsubsection{Disease subtypes discovery}
\label{sec:DT-subtype}
We identify disease subtypes 
using the hierarchical clustering algorithm 
(described in Section~\ref{sec:hete}). 
The CRS values of the top 12 largest network communities in $G_{\text{coSel}}^{\text{DT}}$
are used to describe and represent diseased individuals in the validation cohort. 
We chose the decision tree algorithm over the logistic regression and random forest algorithms 
to generate CRS values.
This ensures the consistency between the CRS generation and feature selection algorithms. 
The CRS values generated using different machine learning algorithms are similar to each other.

\begin{figure}
     \centering
     \begin{subfigure}{0.45\textwidth}
         \centering
         \includegraphics[width=\textwidth]{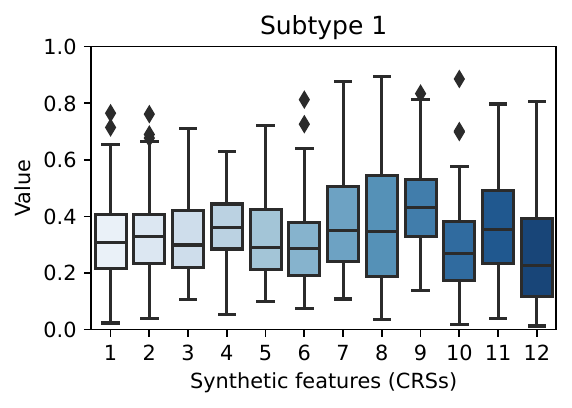}
     \end{subfigure}
     \begin{subfigure}{0.45\textwidth}
         \centering
         \includegraphics[width=\textwidth]{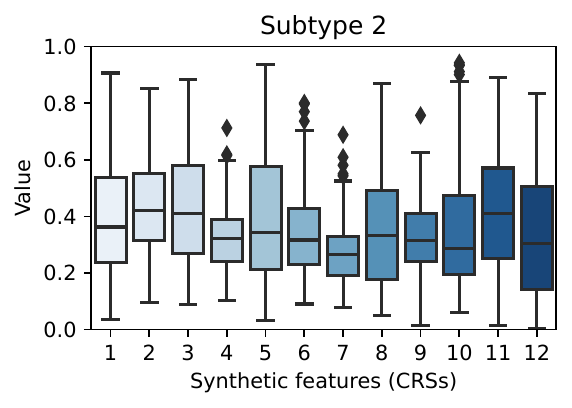}
     \end{subfigure}
     \begin{subfigure}{0.45\textwidth}
         \centering
         \includegraphics[width=\textwidth]{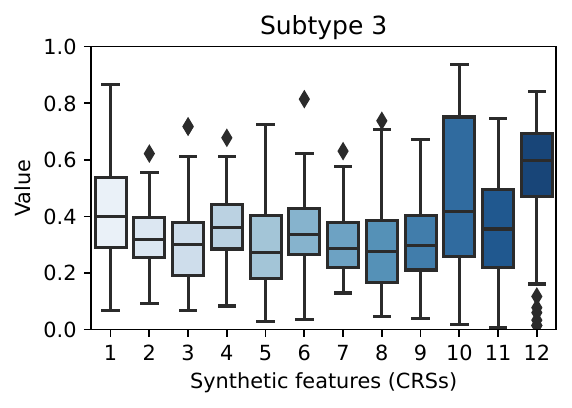}
     \end{subfigure}
     \begin{subfigure}{0.45\textwidth}
         \centering
         \includegraphics[width=\textwidth]{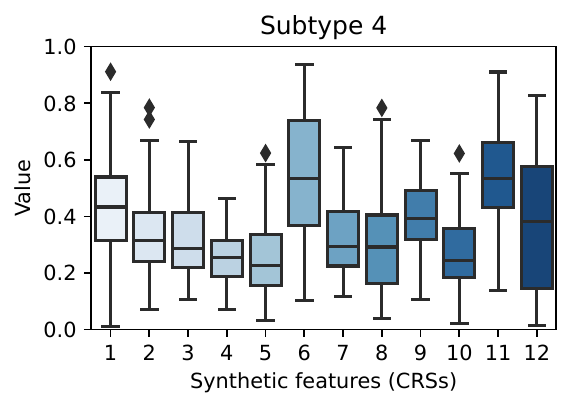}
     \end{subfigure}
    \caption[Cluster means plots for disease subtypes derived from CRSs based on decision tree]{Cluster means plots for disease subtypes derived from CRSs based on decision tree.}
  \label{fig:dtfr_cluster_means}
\end{figure}

The community-specific predictive models 
are trained on the training cohort 
and used to predict the individuals in the validation cohort. 
We then use Euclidean distance and Ward.D linkage 
for heterogeneity analysis on the validation cohort. 
The results suggest that 
the diseased individuals in the validation cohort 
can be divided into four subtypes (see Figure~\ref{fig:dtfr_cluster_means}). 
Subtype one does not have a CRS value with a median value higher than 0.5. 
The CRS with the highest risk is C9 (median=0.442), and the median of the rest of the CRSs is below 0.4. 
Subtype two also does not have a CRS value with a median value higher than 0.5. 
The CRS values with the highest risk are C3 (median=0.411) and C11 (median=0.409). 
The medians of C2 (median=0.421), C3 (median=0.411) and C5 (median=0.343) are higher than the other three subtypes. 
C12 (median=0.596) is at high risk in subtype three.
Subtype four has a high risk in C6 (median=0.532) and C11 (median=0.534).

\subsubsection{Functional enrichment analysis}

\begin{figure}
    \centering
    \includegraphics[width=.8\textwidth]{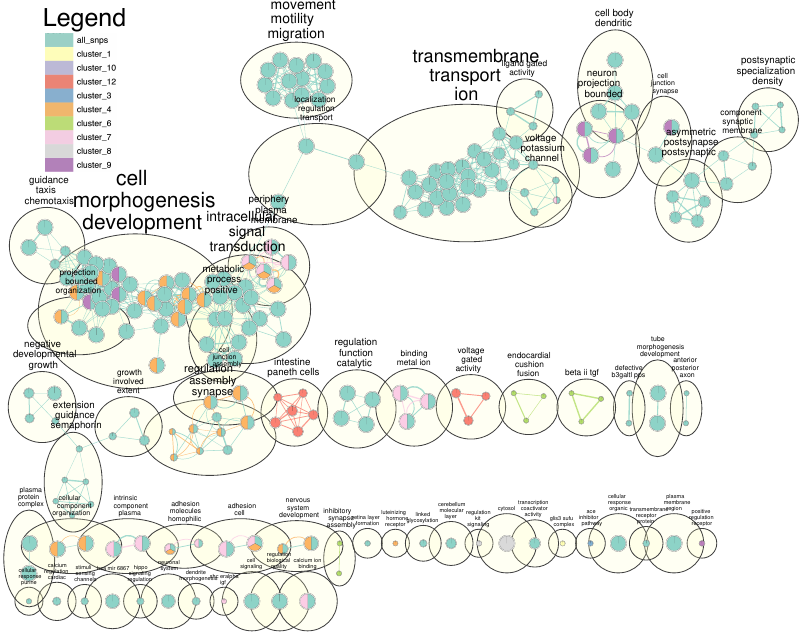}
    \caption{The functional enrichment analysis for $G_{\text{coSel}}^{\text{DT}}$ with $\tau_{\text{occ}}=5$ and $\tau_{\text{cos}}=0.09$. The functional terms (nodes) from different queries are represented by colors.}
    \label{fig:dtfr_enrichment}
\end{figure}

We perform functional enrichment analysis on the 1036 genetic variables 
using g:Profiler~\cite{gProfiler2019} and EnrichmentMap~\cite{Merico2010} 
to identify the theme of the overall variables as well as each variable community. 
We submit a list of 1036 SNP rsIDs to g:Profiler 
and identify 210 biological terms with a $p$-value less than 0.05.
The most significant term is \textit{integral component of plasma membrane} (GO:0005887), 
and the theme of the most significant terms is \textit{cell membrane}. 
We then visualize these terms using the EnrichmentMap tool for network visualization (Figure~\ref{fig:dtfr_enrichment}). 
We also perform enrichment analysis for genetic variables 
in each network community of $G_{\text{coSel}}^{\text{DT}}$ (distinguished by color). 
Communities 6 and 12 contain 6 and 9 terms that are not covered by the enrichment of the entire network.

The resulting network is annotated using the AutoAnnotate tool~\cite{Kucera2016}, 
which identifies groups of similar terms 
according to the network structure 
and determines the theme of each group 
by text-mining the names of biological terms. 
The largest term groups in the network 
are named \textit{cell morphogenesis development} (containing 31 terms), 
\textit{transmembrane transport ion} (containing 25 terms), 
\textit{movement motility migration} (containing 13 terms), 
and \textit{intracellular signal transduction} (containing 11 terms). 
There are also functional differences between different communities of $G_{\text{coSel}}^{\text{DT}}$. 
The genetic variables in Cluster 12 contain a theme 
covering six terms about paneth cell, 
while Cluster 4 linked to terms mainly distribute in \textit{cell morphogenesis development}, 
\textit{regulation assembly synapse}, 
and \textit{periphery plasma membrane}. 
For more details about the enrichment analysis, 
please refer to Supplementary material {\gaheteSeight}.

\subsection{Heterogeneity analysis based on logistic regression}
\label{sec:GA_LR}

This section provides a summary of the heterogeneity analysis 
based on feature selection runs using logistic regression as the fitness measure for the GA. 
The parameter configurations used for this analysis 
are described in Table~\ref{tab:parameter_config}. 
Unlike the decision tree algorithm, 
the logistic regression algorithm focuses on 
discovering genetic variables with high main effects. 
In this section, we examine the risk association of each synthetic feature or CRS value, 
and the disease heterogeneity that they reveal together. 
Additionally, we conduct a genetic enrichment analysis 
to identify the biological terms associated with colorectal cancer.

\begin{figure}[]
     \centering
     \begin{subfigure}[b]{0.45\textwidth}
         \centering
         \includegraphics[width=\textwidth]{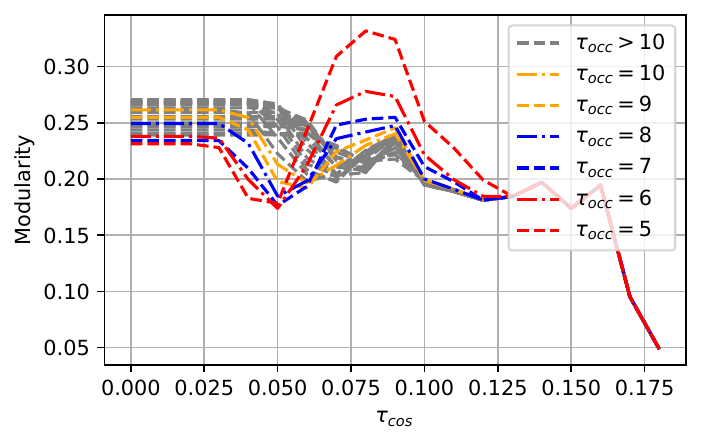}
         \caption{The evolution of network modularity.}
         \label{fig:softthreshold}
     \end{subfigure}
     \hfill
     \begin{subfigure}[b]{0.45\textwidth}
         \centering
         \includegraphics[width=\textwidth]{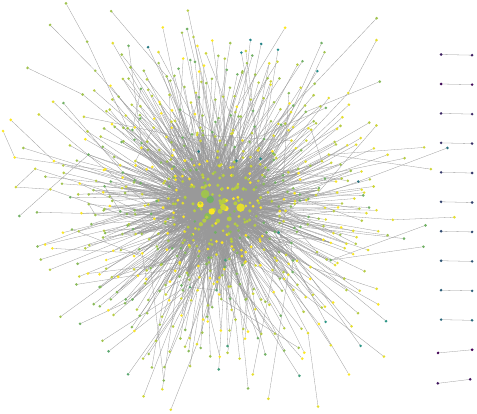}
         \caption{The generated $G_{\text{coSel}}^{\text{LR}}$ network.}
         \label{fig:cosel}
     \end{subfigure}
        \caption[The determination and visualization of feature co-selection network based on logistic regression]{The determination and visualization of feature co-selection network based on logistic regression. We determine the $G_{\text{coSel}}^{\text{LR}}$ network with the highest network modularity based on edge cut-offs $\tau_{\text{occ}}$ and $\tau_{\text{cos}}$. Network communities (N=26) of the resulting network are represented by colors and the size of node represents its degree.}
        \label{fig:two graphs}
\end{figure}

We created a feature co-selection network $G^{\text{LR}}_{\text{coSel}}$ 
by using a set of high-performing feature subsets $\Gamma$ 
produced from approximately 10,000 GA runs. 
In this network, genetic variables are represented as nodes, 
and two types of weights, frequency of co-occurrences and cosine similarity, 
determine the edge between two nodes. 
The edge-cutoff values $\tau_{\text{occ}}$ and $\tau_{\text{cos}}$ 
are determined by optimizing the network modularity (see Figure~\ref{fig:cosel}). 
The network with $\tau_{\text{occ}}=7$ and $\tau_{\text{cos}}=0.08$ 
yields the most suitable $G^{\text{LR}}_{\text{coSel}}$ network (modularity=0.253) 
for further analysis (visualized in Figure~\ref{fig:cosel}). 
Some thresholds that could potentially yield better modularity are excluded 
as they resulted in an excessive number of network components 
or contained too many edges and nodes. 
We consider these networks not suitable for subsequent analysis. 
Detailed network investigation 
using different edge cut-offs 
is provided in the supplementary materials ({\gaheteStwo}). 
The resulting $G^{\text{LR}}_{\text{coSel}}$ network 
has 26 communities of genetic variables, 
with seven communities containing more than ten genetic variables. 
The network communities 
are visualized in Figure~\ref{fig:cosel} using different colors 
and are used to create the synthetic features, as defined in Section~\ref{sec:featueSyn}.

\begin{figure}
     \centering
     \begin{subfigure}{0.19\textwidth}
         \centering
         \includegraphics[width=\textwidth]{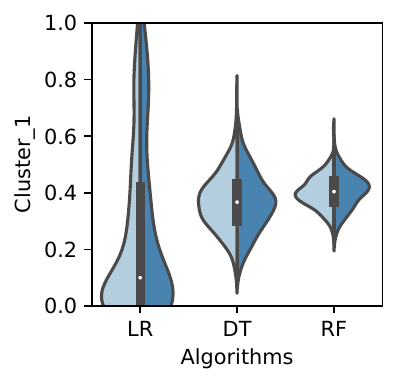}
     \end{subfigure}
     \begin{subfigure}{0.19\textwidth}
         \centering
         \includegraphics[width=\textwidth]{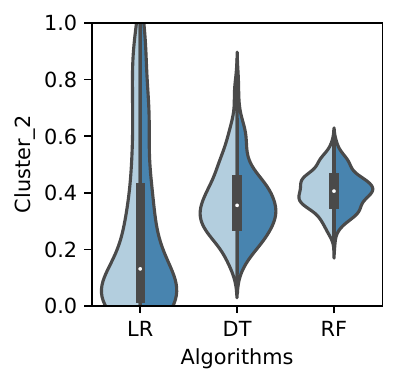}
     \end{subfigure}
     \begin{subfigure}{0.19\textwidth}
         \centering
         \includegraphics[width=\textwidth]{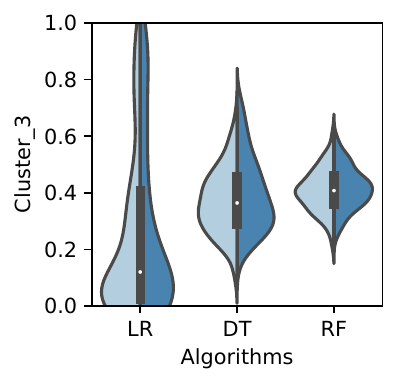}
     \end{subfigure}
     \begin{subfigure}{0.19\textwidth}
         \centering
         \includegraphics[width=\textwidth]{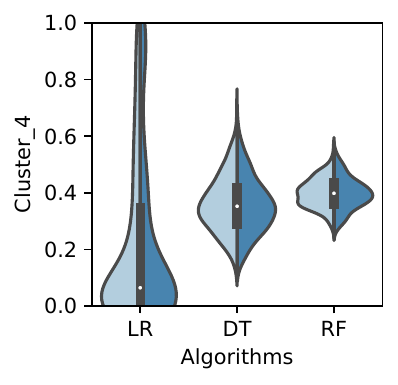}
     \end{subfigure}
     \begin{subfigure}{0.19\textwidth}
         \centering
         \includegraphics[width=\textwidth]{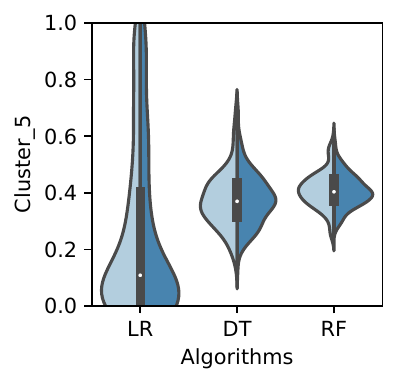}
     \end{subfigure}
     \begin{subfigure}{0.19\textwidth}
         \centering
         \includegraphics[width=\textwidth]{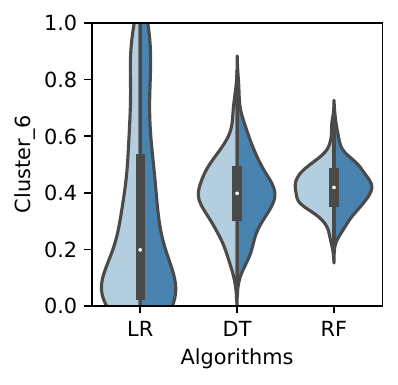}
     \end{subfigure}
     \begin{subfigure}{0.19\textwidth}
         \centering
         \includegraphics[width=\textwidth]{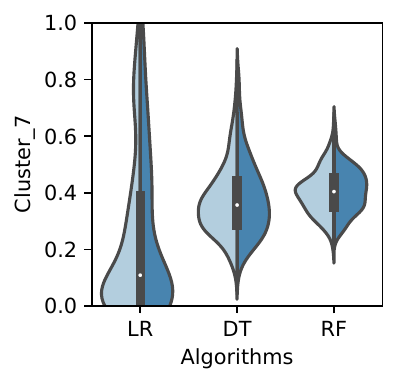}
     \end{subfigure}
     \begin{subfigure}{0.19\textwidth}
         \centering
         \includegraphics[width=\textwidth]{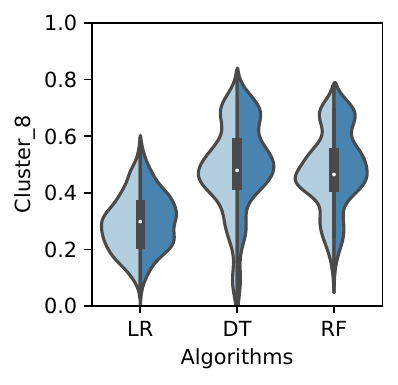}
     \end{subfigure}
     \begin{subfigure}{0.19\textwidth}
         \centering
         \includegraphics[width=\textwidth]{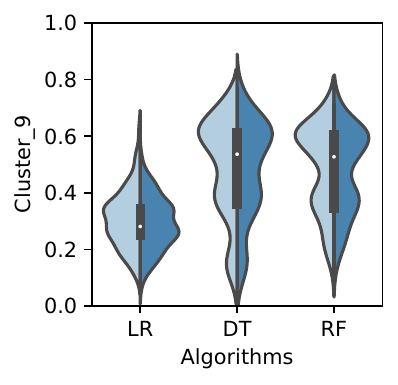}
     \end{subfigure}
     \begin{subfigure}{0.19\textwidth}
         \centering
         \includegraphics[width=\textwidth]{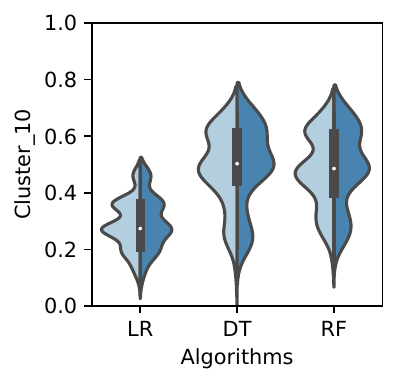}
     \end{subfigure}
     \begin{subfigure}{0.19\textwidth}
         \centering
         \includegraphics[width=\textwidth]{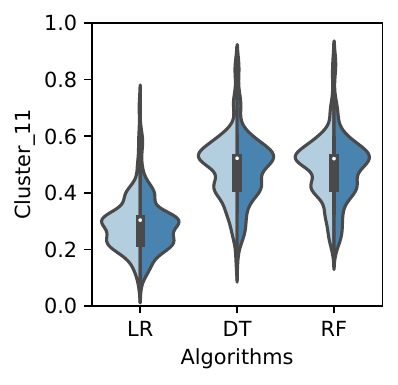}
     \end{subfigure}
     \begin{subfigure}{0.19\textwidth}
         \centering
         \includegraphics[width=\textwidth]{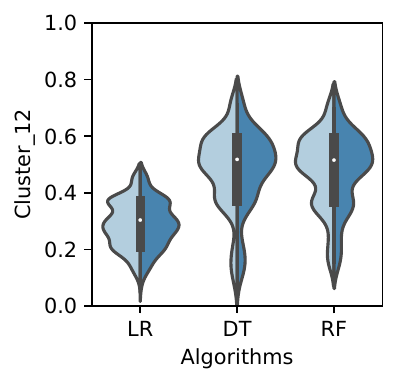}
     \end{subfigure}
     \begin{subfigure}{0.19\textwidth}
         \centering
         \includegraphics[width=\textwidth]{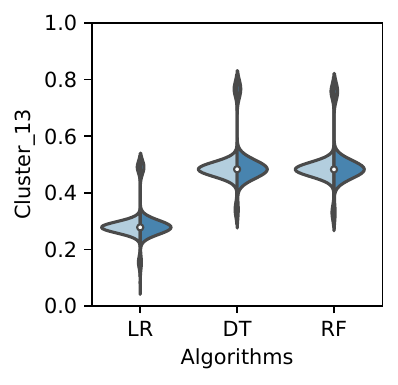}
     \end{subfigure}
     \begin{subfigure}{0.19\textwidth}
         \centering
         \includegraphics[width=\textwidth]{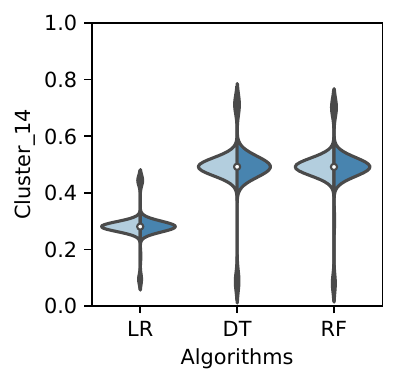}
     \end{subfigure}
     \begin{subfigure}{0.19\textwidth}
         \centering
         \includegraphics[width=\textwidth]{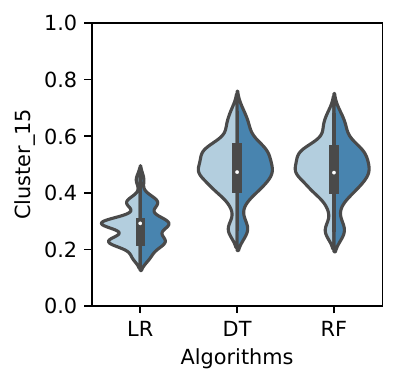}
     \end{subfigure}
     \begin{subfigure}{0.19\textwidth}
         \centering
         \includegraphics[width=\textwidth]{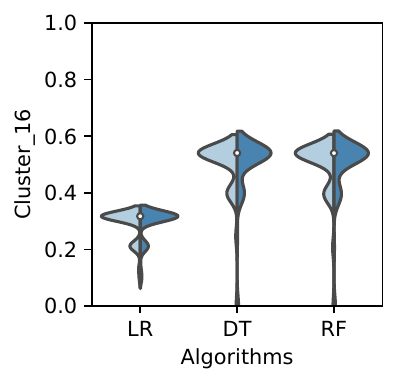}
     \end{subfigure}
     \begin{subfigure}{0.19\textwidth}
         \centering
         \includegraphics[width=\textwidth]{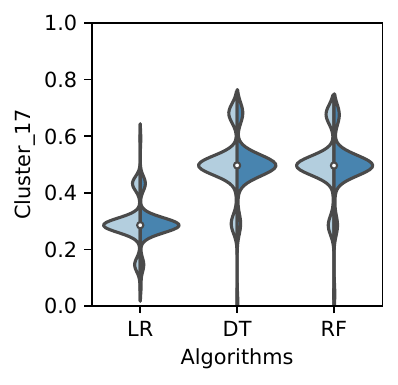}
     \end{subfigure}
     \begin{subfigure}{0.19\textwidth}
         \centering
         \includegraphics[width=\textwidth]{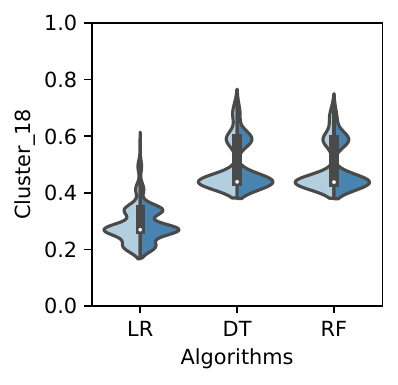}
     \end{subfigure}
     \begin{subfigure}{0.19\textwidth}
         \centering
         \includegraphics[width=\textwidth]{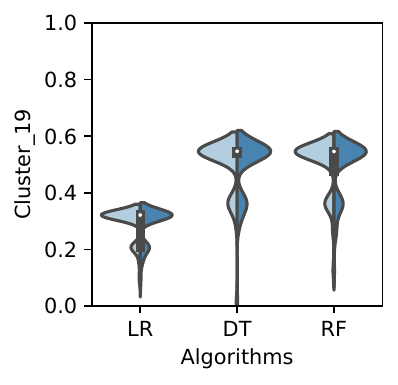}
     \end{subfigure}
     \begin{subfigure}{0.19\textwidth}
         \centering
         \includegraphics[width=\textwidth]{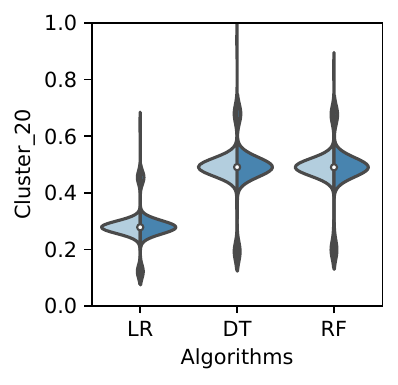}
     \end{subfigure}
     \begin{subfigure}{0.19\textwidth}
         \centering
         \includegraphics[width=\textwidth]{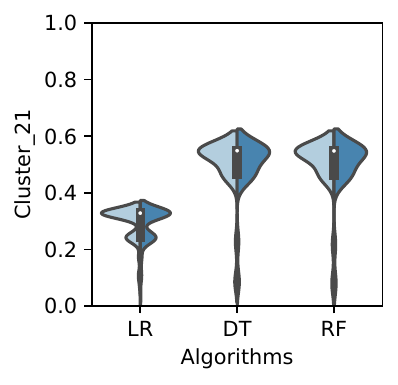}
     \end{subfigure}
     \begin{subfigure}{0.19\textwidth}
         \centering
         \includegraphics[width=\textwidth]{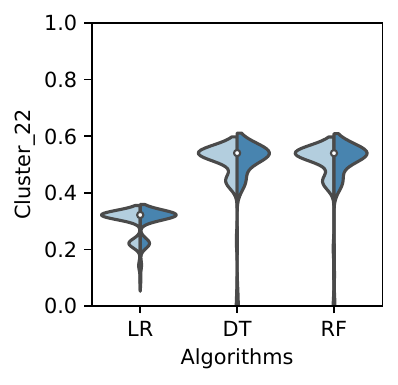}
     \end{subfigure}
     \begin{subfigure}{0.19\textwidth}
         \centering
         \includegraphics[width=\textwidth]{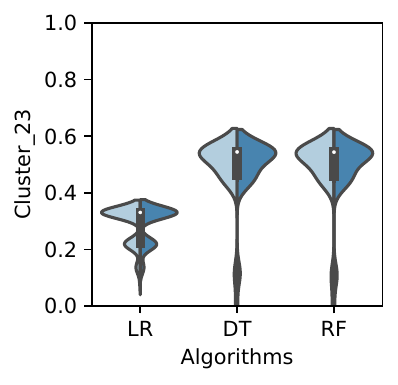}
     \end{subfigure}
     \begin{subfigure}{0.19\textwidth}
         \centering
         \includegraphics[width=\textwidth]{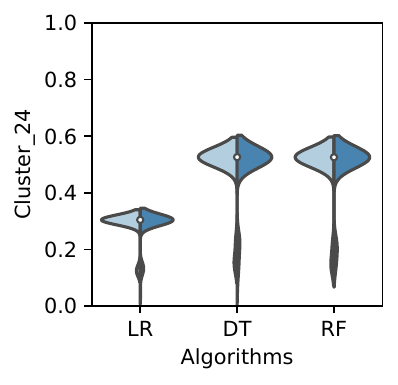}
     \end{subfigure}
     \begin{subfigure}{0.19\textwidth}
         \centering
         \includegraphics[width=\textwidth]{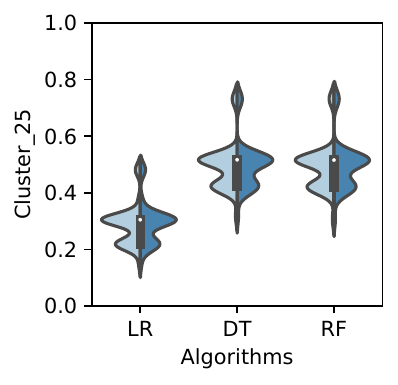}
     \end{subfigure}
     \begin{subfigure}{0.19\textwidth}
         \centering
         \includegraphics[width=\textwidth]{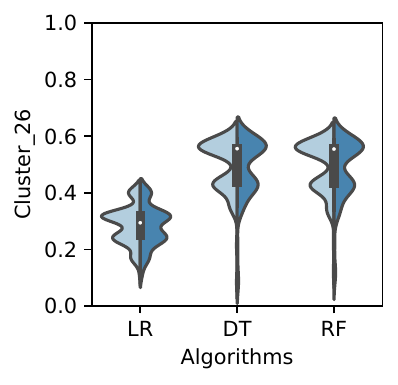}
     \end{subfigure}
     \begin{subfigure}{0.19\textwidth}
         \centering
         \includegraphics[width=\textwidth]{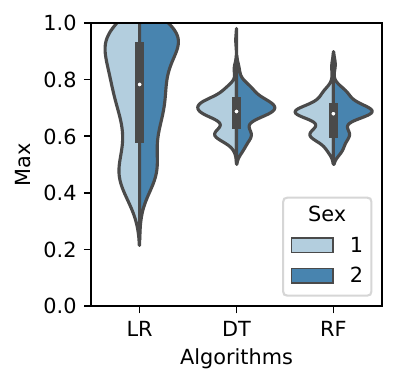}
     \end{subfigure}
        \caption[The heterogeneous risk predictive performance of the CRSs based on logistic regression]{The heterogeneous risk predictive performance of the CRSs based on logistic regression. As many as 26 CRSs communities are identified from $G^{\text{LR}}_{\text{coSel}}$ on the validation cohort. The maximum risk impact of all communities is summarized in sub-figure entitled ``Max''.}
        \label{fig:hete_Pred_LR}
\end{figure}

We evaluate the predictive performance of each synthetic feature (CRS) 
using the resampling method described in Section~\ref{sec:DTFR_Hete_CRS}. 
For each individual in the validation cohort, 
the CRS values are calculated as the mean across 1000 resampling runs. 
Figure~\ref{fig:hete_Pred_LR} explains 
the heterogeneous risk-capturing capability of different CRS values, 
with each individual CRS only capturing the disease risk of a subset of individuals in the validation cohort. 
Notably, the decision tree-based algorithms (decision tree and random forest) 
generally outperform the linear regression algorithms in capturing disease risk. 
However, when we consider the maximum risk captured by all feature subsets, 
the linear regression algorithm performs the best.

\begin{figure}
     \centering
     \begin{subfigure}{0.45\textwidth}
         \centering
         \includegraphics[width=\textwidth]{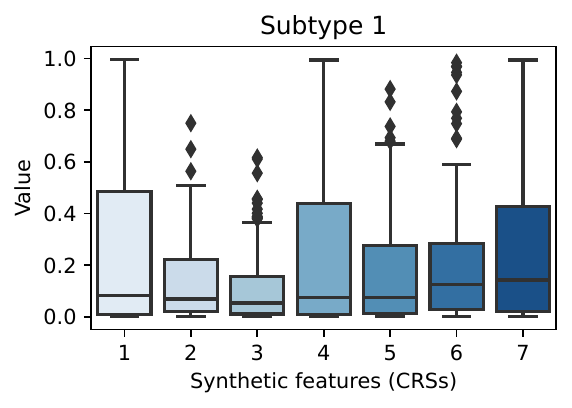}
     \end{subfigure}
     \begin{subfigure}{0.45\textwidth}
         \centering
         \includegraphics[width=\textwidth]{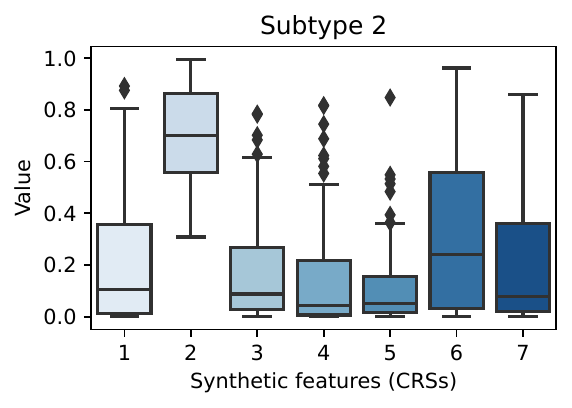}
     \end{subfigure}
     \begin{subfigure}{0.45\textwidth}
         \centering
         \includegraphics[width=\textwidth]{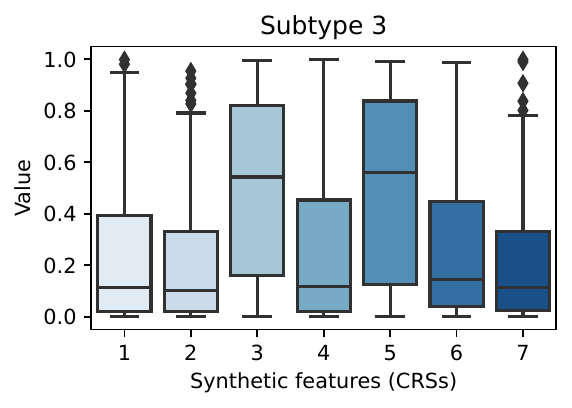}
     \end{subfigure}
     \begin{subfigure}{0.45\textwidth}
         \centering
         \includegraphics[width=\textwidth]{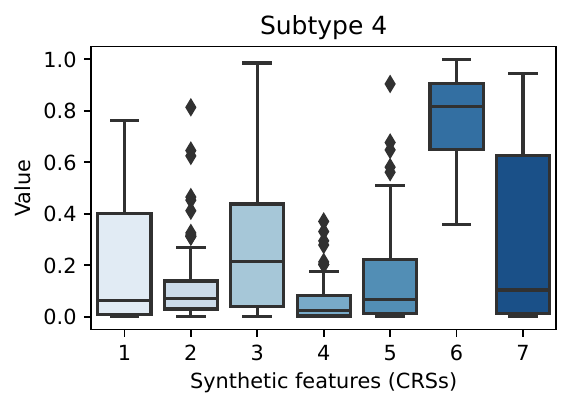}
     \end{subfigure}
        \caption[Cluster means plots for disease subtypes derived from CRSs based on logistic regression]{Cluster means plots for disease subtypes derived from CRSs based on logistic regression.}
        \label{fig:lr_cluster_means}
\end{figure}

We employ seven logistic regression-based CRSs, 
each consisting of more than ten genetic variables, 
to predict and represent the diseased individuals in the validation cohort for the analysis. 
Using hierarchical clustering algorithm 
based on Euclidean distance and Ward.D linkage, 
we identify four disease subtypes as shown in Figure~\ref{fig:lr_cluster_means}. 
Among these subtypes, three contain at least one high-risk CRS with a median greater than 0.5.
Subtype 1 does not have any CRS with a median higher than 0.5. The CRS with the highest risk is C1 (median=0.275), C4 (median=0.252) and C7 (median=0.259), and the median of the rest of the CRSs is below 0.2.
Subtype 2 is at high risk in C2 (median=0.700). The remaining six CRSs present lower risk with a median of less than 0.2.
Subtype 3 is at high risk in C3 (median=0.543) and C5 (median=0.561). The medians of all remaining five CRSs are below 0.2.
Subtype 4 is at high risk in C6 (median=0.815). The remaining six CRSs present lower risk, except from C3 (median=0.214), all with a median less than 0.2.

\begin{figure}
    \centering
    \includegraphics[width=.85\textwidth]{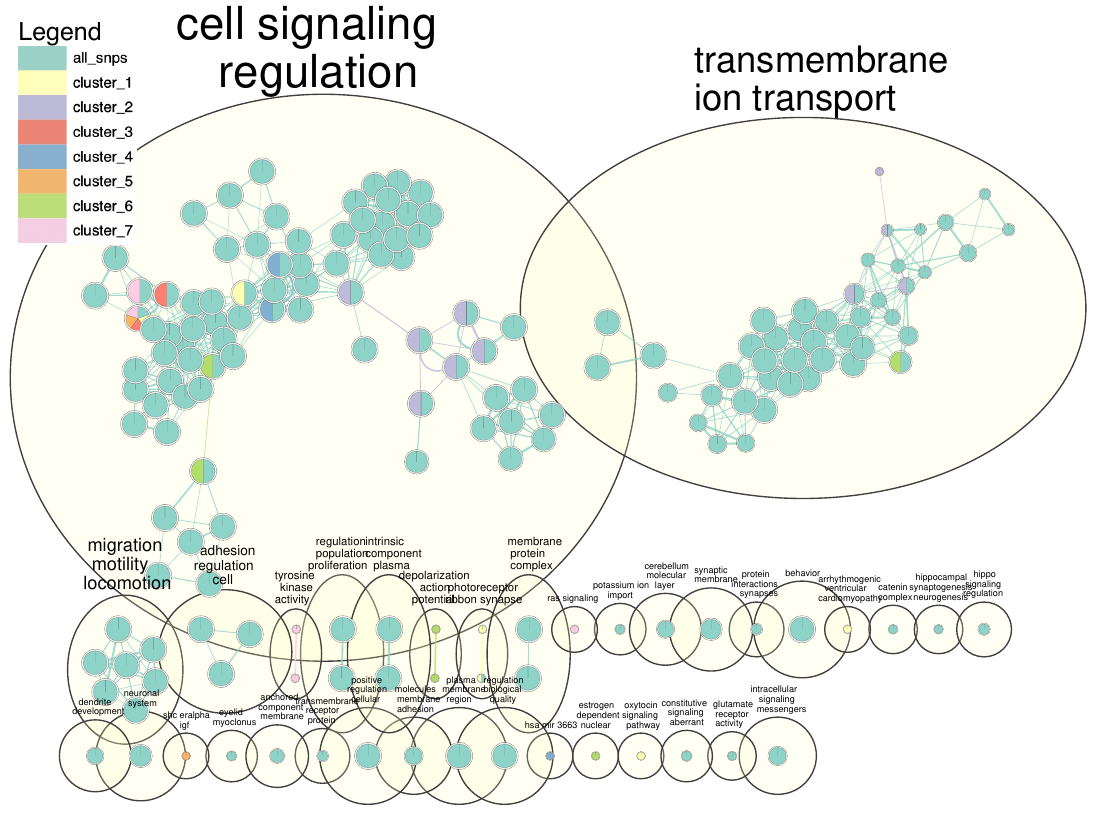}
    \caption{The functional enrichment analysis for $G_{\text{coSel}}^{\text{LR}}$ with $\tau_{\text{occ}}=7$ and $\tau_{\text{cos}}=0.08$. The functional terms (nodes) from different queries are represented by colors.}
    \label{fig:lr_enrichment}
\end{figure}

We submitted a total of 820 genetic variables in $G_{\text{coSel}}^{\text{LR}}$ 
as well as seven community specific variable subsets
for functional enrichment analysis using g:Profiler (see Supplementary materials {\gaheteSnine} for details). 
As shown in Figure~\ref{fig:lr_enrichment}, 
the functional biological terms consist of two main groups. 
The theme of the largest group (N=78) 
is \textit{cell signaling regulation} 
and the theme of the second largest group (N=44) is \textit{transmembrane ion transport}.

\subsection{Over-fitting analysis}

Since the number of observations in our dataset is relatively small 
compared to the number of features, 
we use all healthy controls in the training cohort 
for feature selection. 
This procedure may result in concerns 
regarding the over-fitting issue. 
Therefore, we develop new experiment 
to investigate the predictive performance of the proposed method 
for unseen observations.

We divide the entire dataset (including the training and validation cohorts) 
into training (80\%) and testing (20\%) splits. 
The training dataset is used as the input data for feature selection 
and to generate CRS values describing the observations in the test dataset. 
We then use the student t-test and AUC-ROC 
to investigate the ability of individual CRS values 
in discriminating between diseased and healthy individuals 
in the testing dataset.

We replicate the analysis for different fitness evaluation algorithms for the GA, 
namely the decision tree and logistic regression algorithms, 
and different network thresholds $(\tau_{\text{occ}},\tau_{\text{cos}})\in\{(3,0.12),(4,0.1),(4,0.08)\}$ 
for decision tree-based GA and $(\tau_{\text{occ}},\tau_{\text{cos}})\in\{(3,0.12),(38,0.08),(55,0.09)\}$ 
for logistic regression-based GA to determine the degree of overfitting. 

The results indicate that for all parameter configurations, 
the CRS values containing more than 30 genetic variables 
will have significant ($p<0.05$) discriminative capability 
on the testing dataset (see {\gaheteSthree}, {\gaheteSfour}, {\gaheteSfive}, {\gaheteSsix}, and {\gaheteSseven}). 
Please refer to the Supplementary materials ({\gaheteSten} and {\gaheteSeleven}) 
for detailed predictive performance 
quantified by AUC-ROC on the testing dataset. 
We conclude that both fitness algorithms have good resistance to overfitting.

\section{Discussion}

We propose a novel feature selection framework, named FCS-Net, for GWAS, 
based on the feature co-selection network, $G_{\text{coSel}}$. 
This network facilitates the depiction of collaborative interactions among genetic variables. 
It can be used to discern communities of genetic variables 
contributing to disease heterogeneity. 
By offering a more intricate understanding of the relationships between genetic variables, 
FCS-Net provides a more detailed understanding than
the traditional rank-based feature selection methods usually employed in GWAS.

\subsection{The architecture of coSel network}

The feature co-selection network, $G_{\text{coSel}}$, 
is formulated from a set of evolved high-performing genetic variable subsets. 
Within this network, 
nodes represent genetic variables, 
and the edges connecting pairs of variables 
are weighted by two distinct metrics: 
the frequency of co-occurrence between a pair of variables, 
and the cosine similarity. 
Clustering algorithms are employed 
to detect communities of genetic variables. 
The segmentation of variable subsets 
serves to create synthetic features or Community Risk Scores (CRSs), 
which are subsequently used to evaluate individual disease risk heterogeneity.

Our feature selection algorithm based on GA 
is guided to select feature subsets 
with varying characteristics 
using different fitness measures. 
Our results show that 
the co-selection network generated 
using the decision tree algorithm as fitness evlauation emasure, $G_{\text{coSel}}^{\text{DT}}$ (Figure~\ref{fig:coSel_GADT}), 
differs considerably from the network 
generated using the linear regression algorithm, $G_{\text{coSel}}^{\text{LR}}$ (Figure~\ref{fig:cosel}).

The most significant difference 
between the $G_{\text{coSel}}^{\text{DT}}$ and $G_{\text{coSel}}^{\text{LR}}$ networks 
lies in their structural configurations. 
The $G_{\text{coSel}}^{\text{LR}}$ network 
features a central-peripheral architecture, 
while the $G_{\text{coSel}}^{\text{DT}}$ network 
exhibits a multi-centered architecture. 
This difference is attributed to the distinct mechanisms 
inherent to the two fitness measure algorithms. 
Linear regression-based logistic regression algorithms 
tend to select genetic variables with higher main effects. 
Conversely, the decision tree algorithm 
captures variables with feature interactions 
in addition to those with strong main effects. 
We consider that the multi-centered structure 
more accurately describes the structural modularity 
of the genetic architecture of Colorectal Cancer (CRC) 
than the central-peripheral architecture.

Another significant discrepancy between these two networks 
is on the edge weights. 
In this study, we employ two types of weights: 
the weight based on the frequency of co-occurrence, 
and the cosine similarity distance. 
For the $G_{\text{coSel}}^{\text{DT}}$ network, 
an elevated cosine similarity threshold considerably 
boosts the network's modularity metric, 
from roughly 0.2 to 0.6. 
The $G_{\text{coSel}}^{\text{LR}}$ network, in contrast, 
lacks this property, exhibiting no modularity response 
to an increased cosine similarity threshold. 
The predictive mechanisms' distinctness 
between logistic regression and decision tree algorithms 
contributes this difference. 
The decision tree algorithm 
can detect feature interactions, 
requiring features with such interactions 
to be selected simultaneously to improve the fitness value of a feature subset in the GA. 
Whereas, the logistic regression algorithm 
lacks such a mechanism, 
resulting in the evolved feature subsets 
produced by the GA not reflecting feature dependencies. 
The cosine distance used in our study plays a critical role 
in the construction of the $G_{\text{coSel}}$ network. 
Unlike the commonly used threshold 
based on the number of co-occurrences~\cite{Miani2022}, 
the cosine distance excludes edges 
that exist between frequently selected features, 
allowing the co-selection network's edges 
to reflect the interplays between features.

We consider $G_{\text{coSel}}^{\text{DT}}$ 
more suitable for heterogeneity analysis than $G_{\text{coSel}}^{\text{LR}}$. 
This observation is based on our simulation study, 
which confirm that the decision tree algorithm 
enables the genetic algorithm in capturing feature interactions. 
Therefore, using the decision tree algorithm as the fitness measure of the GA 
can help uncover more disease-associated interaction genetic variables
than the logistic regression algorithm does. 
Furthermore, the multi-centered network structure of $G_{\text{coSel}}^{\text{DT}}$ 
facilitates a more precise identification of heterogeneous subsets of genetic variables 
than the central-peripheral structure.

\subsection{Heterogeneity analysis}
In Section~\ref{sec:synData_analysis}, 
we suggest that the decision tree-based fitness measure 
is preferred than logistic regression 
when dealing with high-dimensional datasets 
with feature interactions. 
Therefore, we replicate GA-based feature selection in this study, 
using both logistic regression and decision tree-based fitness measures. 
This approach allow us to utilize different machine learning techniques 
and discover features that influence disease risk through varying genetic mechanisms.

Our heterogeneity analysis successfully identifies disease subtypes 
in the validation cohort 
that exhibited heterogeneous risk across distinct CRS values, as represented in Figures~\ref{fig:dtfr_cluster_means} and~\ref{fig:lr_cluster_means}. 
However, there is no correlation ($p=0.622$)
between subtypes generated based on different fitness evaluation algorithms. 
Moreover, no significant association is found between disease subtypes and gender.

\subsection{Biological enrichment analysis}

The biological enrichment analysis in this work 
utilizes the procedure documented in \cite{reimand2019pathway}.
This protocol employs network science and natural language processing methods 
to extract themes related to CRC risk 
from a wide range of similar biological terms. 
In this section, we explore and compare the results of the biological enrichment analysis 
produced from the rsIDs in different coSel networks.

The $G_{\text{coSel}}^{\text{DT}}$ network 
contains 1036 genetic variables. 
The enrichment analysis of these 1036 rsIDs using g:profiler 
produces 210 significant ($p<0.05$) terms. 
These biological terms have four major themes: 
\textit{cell morphogenesis development} (N=31), 
\textit{transmembrane ion transport} (N=25), 
\textit{movement motility migration} (N=13), 
and \textit{intracellular signal transduction} (N=11).
To delve further into the role of these themes in CRC, 
we search for content in the NIH Library 
containing both the theme and CRC over the last five years. 
A total of 73 articles are found for \textit{cell morphogenesis development}, 
7 articles for \textit{transmembrane ion transport}, 
144 articles for \textit{movement motility migration}, 
and 384 articles for \textit{intracellular signal transduction}.

The $G_{\text{coSel}}^{\text{LR}}$ network 
contains 820 genetic variables. 
The enrichment analysis of these 820 rsIDs using g:profiler 
produces 171 significant ($p<0.05$) terms.
These biological terms have two major themes: 
\textit{cell signaling regulation} (N=78) and 
\textit{transmembrane ion transport} (N=44). 
We find a total of 5688 articles for cell signaling regulation 
and CRC 
and 7 articles for \textit{transmembrane ion transport} and CRC 
over the last five years, using the NIH Library.

Comparing the enrichment analysis of genetic variables in different networks, 
we can identify the characteristics of the biological functions 
derived from different sets of rsIDs. 
We find that \textit{transmembrane ion transport} 
is a common theme in both selection mechanisms. 
Generally, the $G_{\text{coSel}}^{\text{DT}}$ network generated based on the decision tree 
encompasses more complex biological functional terms, 
which may reflect the multi-centred network architecture of the $G_{\text{coSel}}$ network. 
The biological functions found based on the linear regression model 
have been more widely studied than those found based on the decision tree. 
An explanation might be that individual feature effects 
are still commonly used to select genetic variables for GWAS, 
and the genetic variables associated with feature interaction have not received enough attention in the literature yet.

\section{Conclusions and future works}

The existence of epistasis and heterogeneity 
can render the analysis and interpretation of high-dimensional genetic data challenging. 
This study transforms feature selection in GWAS 
into a combinatorial optimization problem, 
enabling heterogeneity analysis 
in the context of feature interaction. 
A crucial limitation of existing feature selection methods 
is that they only produce a list of features, 
overlooking the collaborative interplay among these features. 
To overcome this, FCS-Net leverages $G_{\text{coSel}}$ 
to identify clusters of interacting variables 
associated with the disease. 
Simulation studies suggests 
that the use of the decision tree algorithm as the fitness evaluation for GA 
helps to recognize feature interactions, 
thereby enhancing the quality of heterogeneity analysis.

We propose several directions for future work.
Firstly, our empirical findings provide new insights 
into feature selection methods based on the wrapper approach. 
We confirm that the search for variables 
with different disease-association mechanisms 
can be achieved by using various machine learning algorithms (e.g., logistic regression and decision tree) 
as the fitness measure of GA. 
As shown in Figures~\ref{fig:coSel_GADT} and~\ref{fig:cosel}, 
the network derived from the decision tree 
has a higher degree of modularity, 
which is undoubtedly beneficial for identifying heterogeneous variable subsets. 
Further research is needed to investigate 
the characteristics and capabilities of different machine learning algorithms 
to enhance the ability of GA or other wrapper feature selection approaches 
to capture disease-related variables.

Secondly, future studies could build upon 
the use of cosine similarity 
for identifying epistatic feature collaborations based on FCS-Net. 
Cosine similarity, as observed in our study, 
serves as an effective metric for epistasis discovery 
when decision tree-based fitness measures are applied. 
Our simulation study suggest a possible connection 
between such feature collaborations and feature interactions. 
This opens up opportunities for future research 
to explore second-order, and even higher-order, feature interactions.
Our initial analysis on synthetic data 
encompassing third-order interactions 
indicates that decision tree-based GA 
can capture these interactions 
and represent them as a clique of three nodes on the feature co-selection network.

Lastly, addressing the limitations of hard community clustering algorithms 
is a crucial aspect for future research. 
Current greedy clustering algorithms segregate nodes into disjoint groups, 
operating under the assumption that a variable can only belong to a single group. 
However, in the field of cancer genomics, 
this assumption may not always hold true. 
The loss of essential variables 
can lead to a drop in the predictive performance of the corresponding CRS. 
Future work could employ network clustering algorithms 
that facilitate overlapping community detection~\cite{n2015overview, blei2003latent, Yang2013}
to overcome this constraint.

\section*{Data availability}
The symulation datasets can be accessed through the Penn Machine Learning Benchmarks~\cite{Romano2021}. 
The GWAS data can be accessed through the colorectal cancer transdisciplinary (CORECT) consortium~\cite{ScOt15}.

\section*{Supplementary materials}
\begin{itemize}
    \item {\gaheteSone}: The evolution of the feature co-selection network based on decision tree by thresholds
    \item {\gaheteStwo}: The evolution of the feature co-selection network based on logistic regression by thresholds
    \item {\gaheteSthree}: Overfitting analysis for feature co-selection network based on decision tree with $\tau_{\text{occ}}=3$ and $\tau_{\text{cos}}=0.12$
    \item {\gaheteSfour}: Overfitting analysis for feature co-selection network based on decision tree with $\tau_{\text{occ}}=4$ and $\tau_{\text{cos}}=0.1$
    \item {\gaheteSfive}: Overfitting analysis for feature co-selection network based on decision tree with $\tau_{\text{occ}}=4$ and $\tau_{\text{cos}}=0.08$
    \item {\gaheteSsix}: Overfitting analysis for feature co-selection network based on logistic regression with $\tau_{\text{occ}}=38$ and $\tau_{\text{cos}}=0.08$
    \item {\gaheteSseven}: Overfitting analysis for feature co-selection network based on logistic regression with $\tau_{\text{occ}}=55$ and $\tau_{\text{cos}}=0.09$
    \item {\gaheteSeight}: Biological enrichment analysis for the feature co-selection network based on decision tree
    \item {\gaheteSnine}: Biological enrichment analysis for the feature co-selection network based on logistic regression
    \item {\gaheteSten}: The AUC-ROC of the CRSs derived from $G_{\text{coSel}}^{DT}$
    \item {\gaheteSeleven}: The AUC-ROC of the CRSs derived from $G_{\text{coSel}}^{LR}$
\end{itemize}

\section*{Acknowledgements}
We are grateful to Digital Research Alliance of Canada and Centre for Advanced Computing at Queen's University for providing high-performance computing infrastructures.

\noindent {\it Conflict of Interest:} No author has competing interests.

\section*{Funding information}
This work was supported by the Natural Sciences and Engineering Research Council (NSERC) of Canada,
Discovery Grant [RGPIN-2023-03302 to T.H.].

\bibliographystyle{unsrt}  
\bibliography{references}  

\begin{thebibliography}{10}

\bibitem{rood2021legacy}
Jennifer~E Rood and Aviv Regev.
\newblock The legacy of the human genome project.
\newblock {\em Science}, 373(6562):1442--1443, 2021.

\bibitem{visscher2021discovery}
Peter~M Visscher, Loic Yengo, Nancy~J Cox, and Naomi~R Wray.
\newblock Discovery and implications of polygenicity of common diseases.
\newblock {\em Science}, 373(6562):1468--1473, 2021.

\bibitem{visscher201710}
Peter~M Visscher, Naomi~R Wray, Qian Zhang, Pamela Sklar, Mark~I McCarthy,
  Matthew~A Brown, and Jian Yang.
\newblock 10 years of {GWAS} discovery: Biology, function, and translation.
\newblock {\em The American Journal of Human Genetics}, 101(1):5--22, 2017.

\bibitem{loos202015}
Ruth~JF Loos.
\newblock 15 years of genome-wide association studies and no signs of slowing
  down.
\newblock {\em Nature Communications}, 11(1):1--3, 2020.

\bibitem{dahl2020genetic}
Andy Dahl and Noah Zaitlen.
\newblock Genetic influences on disease subtypes.
\newblock {\em Annual Review of Genomics and Human Genetics}, 21:413--435,
  2020.

\bibitem{wray2018common}
Naomi~R Wray, Cisca Wijmenga, Patrick~F Sullivan, Jian Yang, and Peter~M
  Visscher.
\newblock Common disease is more complex than implied by the core gene
  omnigenic model.
\newblock {\em Cell}, 173(7):1573--1580, 2018.

\bibitem{urbanowicz2013role}
Ryan~John Urbanowicz, Angeline~S Andrew, Margaret~Rita Karagas, and Jason~H
  Moore.
\newblock Role of genetic heterogeneity and epistasis in bladder cancer
  susceptibility and outcome: a learning classifier system approach.
\newblock {\em Journal of the American Medical Informatics Association},
  20(4):603--612, 2013.

\bibitem{torkamani2018personal}
Ali Torkamani, Nathan~E Wineinger, and Eric~J Topol.
\newblock The personal and clinical utility of polygenic risk scores.
\newblock {\em Nature Reviews Genetics}, 19(9):581--590, 2018.

\bibitem{Gabai-Kapara14205}
Efrat Gabai-Kapara, Amnon Lahad, Bella Kaufman, Eitan Friedman, Shlomo Segev,
  Paul Renbaum, Rachel Beeri, Moran Gal, Julia Grinshpun-Cohen, Karen Djemal,
  Jessica~B. Mandell, Ming~K. Lee, Uziel Beller, Raphael Catane, Mary-Claire
  King, and Ephrat Levy-Lahad.
\newblock Population-based screening for breast and ovarian cancer risk due to
  {\it brca1} and {\it brca2}.
\newblock {\em Proceedings of the National Academy of Sciences},
  111(39):14205--14210, 2014.

\bibitem{ritchie2003power}
Marylyn~D Ritchie, Lance~W Hahn, and Jason~H Moore.
\newblock Power of multifactor dimensionality reduction for detecting gene-gene
  interactions in the presence of genotyping error, missing data, phenocopy,
  and genetic heterogeneity.
\newblock {\em Genetic Epidemiology}, 24(2):150--157, 2003.

\bibitem{urbanowicz2012gametes}
Ryan~J Urbanowicz, Jeff Kiralis, Nicholas~A Sinnott-Armstrong, Tamra Heberling,
  Jonathan~M Fisher, and Jason~H Moore.
\newblock Gametes: a fast, direct algorithm for generating pure, strict,
  epistatic models with random architectures.
\newblock {\em BioData mining}, 5(1):1--14, 2012.

\bibitem{schork200114}
Nicholas~J. Schork, Dani Fallin, Bonnie Thiel, Xiping Xu, Ulrich Broeckel,
  Howard~J. Jacob, and Daniel Cohen.
\newblock The future of genetic case-control studies.
\newblock volume~42 of {\em Advances in Genetics}, pages 191--212. Academic
  Press, 2001.

\bibitem{li2018heterogeneity}
Xiong Li, Liyue Liu, Juan Zhou, and Che Wang.
\newblock Heterogeneity analysis and diagnosis of complex diseases based on
  deep learning method.
\newblock {\em Scientific Reports}, 8(1):1--8, 2018.

\bibitem{terao2019distinct}
Chikashi Terao, Boel Brynedal, Zuomei Chen, Xia Jiang, Helga Westerlind, Monika
  Hansson, Per-Johan Jakobsson, Karin Lundberg, Karl Skriner, Guy Serre, et~al.
\newblock Distinct hla associations with rheumatoid arthritis subsets defined
  by serological subphenotype.
\newblock {\em The American Journal of Human Genetics}, 105(3):616--624, 2019.

\bibitem{hinks2016multidimensional}
Timothy~SC Hinks, Tom Brown, Laurie~CK Lau, Hitasha Rupani, Clair Barber, Scott
  Elliott, Jon~A Ward, Junya Ono, Shoichiro Ohta, Kenji Izuhara, et~al.
\newblock Multidimensional endotyping in patients with severe asthma reveals
  inflammatory heterogeneity in matrix metalloproteinases and chitinase 3--like
  protein 1.
\newblock {\em Journal of Allergy and Clinical Immunology}, 138(1):61--75,
  2016.

\bibitem{li2015identification}
Li~Li, Wei-Yi Cheng, Benjamin~S Glicksberg, Omri Gottesman, Ronald Tamler, Rong
  Chen, Erwin~P Bottinger, and Joel~T Dudley.
\newblock Identification of type 2 diabetes subgroups through topological
  analysis of patient similarity.
\newblock {\em Science Translational Medicine}, 7(311):311ra174--311ra174,
  2015.

\bibitem{nicolau2011topology}
Monica Nicolau, Arnold~J Levine, and Gunnar Carlsson.
\newblock Topology based data analysis identifies a subgroup of breast cancers
  with a unique mutational profile and excellent survival.
\newblock {\em Proceedings of the National Academy of Sciences},
  108(17):7265--7270, 2011.

\bibitem{urbanowicz2010application}
Ryan~J. Urbanowicz and Jason~H. Moore.
\newblock The application of pittsburgh-style learning classifier systems to
  address genetic heterogeneity and epistasis in association studies.
\newblock In Robert Schaefer, Carlos Cotta, Joanna Ko{\l}odziej, and G{\"u}nter
  Rudolph, editors, {\em Parallel Problem Solving from Nature, PPSN XI}, pages
  404--413, Berlin, Heidelberg, 2010. Springer Berlin Heidelberg.

\bibitem{urbanowicz2012instance}
Ryan Urbanowicz, Ambrose Granizo-Mackenzie, and Jason Moore.
\newblock Instance-linked attribute tracking and feedback for michigan-style
  supervised learning classifier systems.
\newblock In {\em Proceedings of the 14th annual conference on Genetic and
  evolutionary computation}, pages 927--934, 2012.

\bibitem{urbanowicz2018attribute}
Ryan~J. Urbanowicz, Christopher Lo, John~H. Holmes, and Jason~H. Moore.
\newblock Attribute tracking: Strategies towards improved detection and
  characterization of complex associations.
\newblock In {\em Proceedings of the Genetic and Evolutionary Computation
  Conference}, GECCO '18, page 553–560, New York, NY, USA, 2018. Association
  for Computing Machinery.

\bibitem{zhang2021lcs}
Robert Zhang, Rachael Stolzenberg-Solomon, Shannon~M Lynch, and Ryan~J
  Urbanowicz.
\newblock Lcs-dive: An automated rule-based machine learning visualization
  pipeline for characterizing complex associations in classification.
\newblock {\em arXiv preprint arXiv:2104.12844}, 2021.

\bibitem{dash1997feature}
Manoranjan Dash and Huan Liu.
\newblock Feature selection for classification.
\newblock {\em Intelligent Data Analysis}, 1(1-4):131--156, 1997.

\bibitem{guyon2003introduction}
Isabelle Guyon and Andr{\'e} Elisseeff.
\newblock An introduction to variable and feature selection.
\newblock {\em Journal of Machine Learning Research}, 3(Mar):1157--1182, 2003.

\bibitem{Liu2009}
Huan Liu and Zheng Zhao.
\newblock {\em Manipulating Data and Dimension Reduction Methods: Feature
  Selection}, pages 5348--5359.
\newblock Springer New York, New York, NY, 2009.

\bibitem{liu2010feature}
Huan Liu, Hiroshi Motoda, Rudy Setiono, and Zheng Zhao.
\newblock Feature selection: An ever evolving frontier in data mining.
\newblock In {\em Proceedings of the Fourth International Workshop on Feature
  Selection in Data Mining}, pages 4--13. Proceedings of Machine Learning
  Research, 2010.

\bibitem{breiman2001random}
Leo Breiman.
\newblock Random forests.
\newblock {\em Machine learning}, 45(1):5--32, 2001.

\bibitem{Mak2017}
Timothy Shin~Heng Mak, Robert~Milan Porsch, Shing~Wan Choi, Xueya Zhou, and
  Pak~Chung Sham.
\newblock Polygenic scores via penalized regression on summary statistics.
\newblock {\em Genetic Epidemiology}, 41(6):469--480, 2017.

\bibitem{kira1992practical}
Kenji Kira and Larry~A Rendell.
\newblock A practical approach to feature selection.
\newblock In {\em Machine Learning Proceedings 1992}, pages 249--256. Elsevier,
  1992.

\bibitem{urbanowicz2018relief}
Ryan~J Urbanowicz, Melissa Meeker, William La~Cava, Randal~S Olson, and Jason~H
  Moore.
\newblock Relief-based feature selection: Introduction and review.
\newblock {\em Journal of Biomedical Informatics}, 85:189--203, 2018.

\bibitem{urbanowicz2018benchmarking}
Ryan~J Urbanowicz, Randal~S Olson, Peter Schmitt, Melissa Meeker, and Jason~H
  Moore.
\newblock Benchmarking relief-based feature selection methods for
  bioinformatics data mining.
\newblock {\em Journal of Biomedical Informatics}, 85:168--188, 2018.

\bibitem{yang2008feature}
Cheng-San Yang, Li-Yeh Chuang, Yu-Jung Chen, and Cheng-Hong Yang.
\newblock Feature selection using memetic algorithms.
\newblock In {\em 2008 Third International Conference on Convergence and Hybrid
  Information Technology}, volume~1, pages 416--423. IEEE, 2008.

\bibitem{vilhjalmsson2015modeling}
Bjarni~J Vilhj{\'a}lmsson, Jian Yang, Hilary~K Finucane, Alexander Gusev, Sara
  Lindstr{\"o}m, Stephan Ripke, Giulio Genovese, Po-Ru Loh, Gaurav Bhatia, Ron
  Do, et~al.
\newblock Modeling linkage disequilibrium increases accuracy of polygenic risk
  scores.
\newblock {\em The American Journal of Human Genetics}, 97(4):576--592, 2015.

\bibitem{siedlecki1993note}
W.~Siedlecki and J.~Sklansky.
\newblock A note on genetic algorithms for large-scale feature selection.
\newblock {\em Pattern Recognition Letters}, 10(5):335--347, 1989.

\bibitem{al2017examining}
Murad Al-Rajab, Joan Lu, and Qiang Xu.
\newblock Examining applying high performance genetic data feature selection
  and classification algorithms for colon cancer diagnosis.
\newblock {\em Computer Methods and Programs in Biomedicine}, 146:11--24, 2017.

\bibitem{swerhun2020summary}
Mekaal Swerhun, Jasmine Foley, Brandon Massop, and Vijay Mago.
\newblock A summary of the prevalence of genetic algorithms in bioinformatics
  from 2015 onwards.
\newblock {\em arXiv preprint arXiv:2008.09017}, 2020.

\bibitem{sayed2019nested}
Sabah Sayed, Mohammad Nassef, Amr Badr, and Ibrahim Farag.
\newblock A nested genetic algorithm for feature selection in high-dimensional
  cancer microarray datasets.
\newblock {\em Expert Systems with Applications}, 121:233--243, 2019.

\bibitem{garcia2020unsupervised}
Pilar Garc{\'\i}a-D{\'\i}az, Isabel S{\'a}nchez-Berriel, Juan~A
  Mart{\'\i}nez-Rojas, and Ana~M Diez-Pascual.
\newblock Unsupervised feature selection algorithm for multiclass cancer
  classification of gene expression rna-seq data.
\newblock {\em Genomics}, 112(2):1916--1925, 2020.

\bibitem{yang_pbmdr_2019}
Cheng-Hong Yang, Huai-Shuo Yang, and Li-Yeh Chuang.
\newblock Pbmdr: A particle swarm optimization-based multifactor dimensionality
  reduction for the detection of multilocus interactions.
\newblock {\em Journal of Theoretical Biology}, 461:68--75, 2019.

\bibitem{Wang2022}
Peng Wang, Bing Xue, Jing Liang, and Mengjie Zhang.
\newblock {D}ifferential {E}volution {B}ased {F}eature {S}election: {A}
  {N}iching-based {M}ulti-objective {A}pproach.
\newblock {\em IEEE Transactions on Evolutionary Computation}, pages 1--1,
  2022.

\bibitem{nurhayati_particle_2020}
Nurhayati, Fajar Agustian, and Muhammad Dzil~Ikram Lubis.
\newblock Particle swarm optimization feature selection for breast cancer
  prediction.
\newblock In {\em 2020 8th International Conference on Cyber and IT Service
  Management (CITSM)}, pages 1--6, 2020.

\bibitem{leardi1992genetic}
Riccardo Leardi, R~Boggia, and M~Terrile.
\newblock Genetic algorithms as a strategy for feature selection.
\newblock {\em Journal of Chemometrics}, 6(5):267--281, 1992.

\bibitem{Sha2021}
Zhendong Sha, Ting Hu, and Yuanzhu Chen.
\newblock Feature selection for polygenic risk scores using genetic algorithm
  and network science.
\newblock In {\em 2021 IEEE Congress on Evolutionary Computation (CEC)}, pages
  802--808, 2021.

\bibitem{da2011improving}
S{\'e}Rgio~Francisco Da~Silva, Marcela~Xavier Ribeiro, Jo{\~a}o do ES~Batista
  Neto, Caetano Traina-Jr, and Agma~JM Traina.
\newblock Improving the ranking quality of medical image retrieval using a
  genetic feature selection method.
\newblock {\em Decision Support Systems}, 51(4):810--820, 2011.

\bibitem{canuto2012genetic}
Anne~MP Canuto and Diego~SC Nascimento.
\newblock A genetic-based approach to features selection for ensembles using a
  hybrid and adaptive fitness function.
\newblock In {\em The 2012 International Joint Conference on Neural Networks
  (IJCNN)}, pages 1--8. IEEE, 2012.

\bibitem{sousa2013email}
PEDRO SOUSA, PAULO CORTEZ, RUI VAZ, MIGUEL ROCHA, and MIGUEL RIO.
\newblock Email spam detection: A symbiotic feature selection approach fostered
  by evolutionary computation.
\newblock {\em International Journal of Information Technology \& Decision
  Making}, 12(04):863--884, 2013.

\bibitem{seo2014feature}
Jae-Hyun Seo, Yong~Hee Lee, and Yong-Hyuk Kim.
\newblock Feature selection for very short-term heavy rainfall prediction using
  evolutionary computation.
\newblock {\em Advances in Meteorology}, 2014, 2014.

\bibitem{winkler2011identification}
Stephan~M Winkler, Michael Affenzeller, Witold Jacak, and Herbert Stekel.
\newblock Identification of cancer diagnosis estimation models using
  evolutionary algorithms: a case study for breast cancer, melanoma, and cancer
  in the respiratory system.
\newblock In {\em Proceedings of the 13th Annual Conference Companion on
  Genetic and Evolutionary Computation}, pages 503--510, 2011.

\bibitem{souza2011co}
Francisco Souza, Tiago Matias, and Rui Araójo.
\newblock Co-evolutionary genetic multilayer perceptron for feature selection
  and model design.
\newblock In {\em ETFA2011}, pages 1--7, 2011.

\bibitem{oreski2014genetic}
Stjepan Oreski and Goran Oreski.
\newblock Genetic algorithm-based heuristic for feature selection in credit
  risk assessment.
\newblock {\em Expert Systems with Applications}, 41(4):2052--2064, 2014.

\bibitem{ScOt15}
Fredrick~R Schumacher et~al.
\newblock Genome-wide association study of colorectal cancer identifies six new
  susceptibility loci.
\newblock {\em Nature Communications}, 6:7138, Jul 2015.

\bibitem{das2016next}
Sayantan Das, Lukas Forer, Sebastian Sch{\"o}nherr, Carlo Sidore, Adam~E Locke,
  Alan Kwong, Scott~I Vrieze, Emily~Y Chew, Shawn Levy, Matt McGue, et~al.
\newblock Next-generation genotype imputation service and methods.
\newblock {\em Nature Genetics}, 48(10):1284--1287, 2016.

\bibitem{PuOt07}
Shaun Purcell, Benjamin Neale, Kathe Todd-Brown, Lori Thomas, Manuel~A.R.
  Ferreira, David Bender, Julian Maller, Pamela Sklar, Paul~I.W. {de Bakker},
  Mark~J. Daly, and Pak~C. Sham.
\newblock Plink: A {T}ool {S}et for {W}hole-{G}enome {A}ssociation and
  {P}opulation-{B}ased {L}inkage {A}nalyses.
\newblock {\em The American Journal of Human Genetics}, 81(3):559--575, 2007.

\bibitem{loh2016reference}
Po-Ru Loh, Petr Danecek, Pier~Francesco Palamara, Christian Fuchsberger, Yakir
  A~Reshef, Hilary K~Finucane, Sebastian Schoenherr, Lukas Forer, Shane
  McCarthy, Goncalo~R Abecasis, et~al.
\newblock {R}eference-based phasing using the {H}aplotype {R}eference
  {C}onsortium panel.
\newblock {\em Nature genetics}, 48(11):1443--1448, 2016.

\bibitem{hanley1982meaning}
James~A Hanley and Barbara~J McNeil.
\newblock The meaning and use of the area under a receiver operating
  characteristic (roc) curve.
\newblock {\em Radiology}, 143(1):29--36, 1982.

\bibitem{scikit-learn}
F.~Pedregosa, G.~Varoquaux, A.~Gramfort, V.~Michel, B.~Thirion, O.~Grisel,
  M.~Blondel, P.~Prettenhofer, R.~Weiss, V.~Dubourg, J.~Vanderplas, A.~Passos,
  D.~Cournapeau, M.~Brucher, M.~Perrot, and E.~Duchesnay.
\newblock Scikit-learn: Machine learning in {P}ython.
\newblock {\em Journal of Machine Learning Research}, 12:2825--2830, 2011.

\bibitem{cox1958regression}
David~R Cox.
\newblock The regression analysis of binary sequences.
\newblock {\em Journal of the Royal Statistical Society: Series B
  (Methodological)}, 20(2):215--232, 1958.

\bibitem{breiman2017classification}
Leo Breiman, Jerome~H Friedman, Richard~A Olshen, and Charles~J Stone.
\newblock {\em Classification and regression trees}.
\newblock Routledge, 2017.

\bibitem{DEAP_JMLR2012}
F\'elix-Antoine Fortin, Fran\c{c}ois-Michel {De Rainville}, Marc-Andr\'e
  Gardner, Marc Parizeau, and Christian Gagn\'e.
\newblock {DEAP}: Evolutionary algorithms made easy.
\newblock {\em Journal of Machine Learning Research}, 13:2171--2175, jul 2012.

\bibitem{clauset2004finding}
Aaron Clauset, Mark~EJ Newman, and Cristopher Moore.
\newblock Finding community structure in very large networks.
\newblock {\em Physical Review E}, 70(6):066111, 2004.

\bibitem{kaufman2009finding}
Leonard Kaufman and Peter~J Rousseeuw.
\newblock {\em Finding Groups in Data: An Introduction to Cluster Analysis},
  volume 344.
\newblock John Wiley \& Sons, 2009.

\bibitem{JASP2019}
{JASP Team}.
\newblock {JASP (Version 0.10.2)[Computer software]}, 2019.

\bibitem{gProfiler2019}
Uku Raudvere, Liis Kolberg, Ivan Kuzmin, Tambet Arak, Priit Adler, Hedi
  Peterson, and Jaak Vilo.
\newblock {g:Profiler: a web server for functional enrichment analysis and
  conversions of gene lists (2019 update)}.
\newblock {\em Nucleic Acids Research}, 47(W1):W191--W198, 05 2019.

\bibitem{reimand2019pathway}
J{\"u}ri Reimand, Ruth Isserlin, Veronique Voisin, Mike Kucera, Christian
  Tannus-Lopes, Asha Rostamianfar, Lina Wadi, Mona Meyer, Jeff Wong, Changjiang
  Xu, et~al.
\newblock {P}athway enrichment analysis and visualization of omics data using
  g: {P}rofiler, {G}{S}{E}{A}, {C}ytoscape and {E}nrichment{M}ap.
\newblock {\em Nature Protocols}, 14(2):482--517, 2019.

\bibitem{Merico2010}
Daniele Merico, Ruth Isserlin, Oliver Stueker, Andrew Emili, and Gary~D. Bader.
\newblock Enrichment map: A network-based method for gene-set enrichment
  visualization and interpretation.
\newblock {\em PLOS ONE}, 5(11):1--12, 11 2010.

\bibitem{Kucera2016}
M~Kucera, R~Isserlin, A~Arkhangorodsky, and GD~Bader.
\newblock Autoannotate: A cytoscape app for summarizing networks with semantic
  annotations [version 1; peer review: 2 approved].
\newblock {\em F1000Research}, 5(1717), 2016.

\bibitem{Romano2021}
Joseph~D Romano, Trang~T Le, William La~Cava, John~T Gregg, Daniel~J Goldberg,
  Praneel Chakraborty, Natasha~L Ray, Daniel Himmelstein, Weixuan Fu, and
  Jason~H Moore.
\newblock {PMLB v1.0: an open-source dataset collection for benchmarking
  machine learning methods}.
\newblock {\em Bioinformatics}, 38(3):878--880, 10 2021.

\bibitem{boyle2017expanded}
Evan~A. Boyle, Yang~I. Li, and Jonathan~K. Pritchard.
\newblock An expanded view of complex traits: From polygenic to omnigenic.
\newblock {\em Cell}, 169(7):1177--1186, 2017.

\bibitem{Miani2022}
Alessandro Miani, Thomas Hills, and Adrian Bangerter.
\newblock Interconnectedness and (in)coherence as a signature of conspiracy
  worldviews.
\newblock {\em Science Advances}, 8(43):eabq3668, 2022.

\bibitem{n2015overview}
Chiheb-Eddine~Ben N’Cir, Guillaume Cleuziou, and Nadia Essoussi.
\newblock Overview of overlapping partitional clustering methods.
\newblock In {\em Partitional Clustering Algorithms}, pages 245--275. Springer,
  2015.

\bibitem{blei2003latent}
David~M Blei, Andrew~Y Ng, and Michael~I Jordan.
\newblock Latent dirichlet allocation.
\newblock {\em Journal of Machine Learning Research}, 3:993--1022, 2003.

\bibitem{Yang2013}
Jaewon Yang and Jure Leskovec.
\newblock Overlapping community detection at scale: A nonnegative matrix
  factorization approach.
\newblock In {\em Proceedings of the Sixth ACM International Conference on Web
  Search and Data Mining}, WSDM '13, page 587–596, New York, NY, USA, 2013.
  Association for Computing Machinery.

\end{thebibliography}

\end{document}